\documentclass{article}

\usepackage[preprint]{neurips_data_2024}

\usepackage[utf8]{inputenc} %
\usepackage[T1]{fontenc}    %
\usepackage{hyperref}       %
\usepackage{url}            %
\usepackage{booktabs}       %
\usepackage{amsfonts}       %
\usepackage{nicefrac}       %
\usepackage{microtype}      %
\usepackage{xcolor}         %

\usepackage{graphicx}
\usepackage{amsmath}
\usepackage{amssymb}
\usepackage{mathtools}
\usepackage[ruled]{algorithm2e}
\usepackage{algorithmic}
\usepackage{multirow} 
\usepackage{wrapfig}
\usepackage{subcaption}
\usepackage{tabularray}
\usepackage{longtable}
\usepackage{pythonhighlight}
\usepackage[scientific-notation=true]{siunitx}

\def\max{\mathop{\rm max}\nolimits}

\title{TabularBench: Benchmarking Adversarial Robustness for Tabular Deep Learning in Real-world Use-cases}

\author{%
  Thibault Simonetto \\
  University of Luxembourg\\
  Luxembourg \\
  \texttt{thibault.simonetto@uni.lu} \\
  \And
  Salah Ghamizi \\
  LIST / RIKEN AIP \\
  Luxembourg \\
  \texttt{salah.ghamizi@gmail.com} \\
  \And
  Maxime Cordy \\
  University of Luxembourg\\
  Luxembourg \\
  \texttt{maxime.cordy@uni.lu} \\
}

\begin{document}

\maketitle

\begin{abstract}
  While adversarial robustness in computer vision is a mature research field, fewer researchers have tackled the evasion attacks against tabular deep learning, and even fewer investigated robustification mechanisms and reliable defenses. We hypothesize that this lag in the research on tabular adversarial attacks is in part due to the lack of standardized benchmarks. 
To fill this gap, we propose TabularBench, the first comprehensive benchmark of robustness of tabular deep learning classification models. We evaluated adversarial robustness with CAA%
, an ensemble of gradient and search attacks which was recently demonstrated as the most effective attack against a tabular model. 
In addition to our open benchmark \url{https://github.com/serval-uni-lu/tabularbench} where we welcome submissions of new models and defenses, we implement 7 robustification mechanisms inspired by state-of-the-art defenses in computer vision and propose the largest benchmark of robust tabular deep learning over 200 models across five critical scenarios in finance, healthcare and security. %
We curated %
real datasets for each use case, augmented with hundreds of thousands of realistic synthetic inputs, and trained and assessed our models with and without data augmentations. %
We open-source our library that provides API access to all our pre-trained robust tabular models, and the largest datasets of real and synthetic tabular inputs.  
Finally, %
we analyze the impact of various defenses on the robustness %
and provide actionable insights to design new defenses and robustification mechanisms.

\end{abstract}

\section{Introduction}

Modern machine learning (ML) models have reached or surpassed human-level performance in numerous tasks, leading to their adoption in critical settings such as finance, security, and healthcare. However, concomitantly to their increasing deployment, researchers have uncovered significant vulnerabilities in generating valid adversarial examples (i.e., constraint-satisfying) where test or deployment data are manipulated to deceive the model. Most analyses of these performance drops have focused on the fields of Computer Vision and Large Language Models where extensive benchmarks for adversarial robustness are available (e.g., \citet{croce2020robustbench} and ~\citet{wang2023decodingtrust}).

Despite the widespread use of tabular data and the maturity of Deep Learning (DL) models for this field, the impact of evasion attacks on tabular data has not been thoroughly investigated. Although there are existing benchmarks for \emph{in-distribution} (ID) tabular classification ~\citep{borisov2021deep}, and distribution shifts ~\citep{gardner2023benchmarking}, there is no available benchmark of adversarial robustness for deep tabular models, in particular in critical real-world settings. We summarize in Table \ref{tab:existing_benchmarks} these related benchmarks.

\begin{table}[!ht]
    \centering
    \caption{Existing related benchmarks and their differences with ours}
    \begin{tabular}{c|c|c|c}
    \toprule
        Benchmark &  Domain &  Metric & Realistic evaluation\\
        \midrule
        Tabsurvey~\citep{borisov2021deep} &  Tabular & ID performance & No \\

        Tableshift~\citep{gardner2023benchmarking} &  Tabular & OOD performance & No \\ %
        
        ARES~\citep{dong2020benchmarking} &  CV & Adversarial performance & No \\ %
        
        Robustbench~\citep{croce2020robustbench} &  CV & Adversarial performance & Yes \\
        DecodingTrust~\citep{wang2023decodingtrust} & LLM & Trust (incl adversarial) & Yes \\
        \midrule
        \textbf{OURS} &  Tabular & Adversarial performance & Yes \\
        \bottomrule
    \end{tabular}
    
    \label{tab:existing_benchmarks}
\end{table}

The need for dedicated benchmarks for tabular model robustness is enhanced by the unique challenges that tabular machine learning raises compared to computer vision and NLP tasks. 

\begin{wrapfigure}{r}{0.5\textwidth}
    \centering
    \vspace{-1em}
    \includegraphics[width=.8\linewidth]{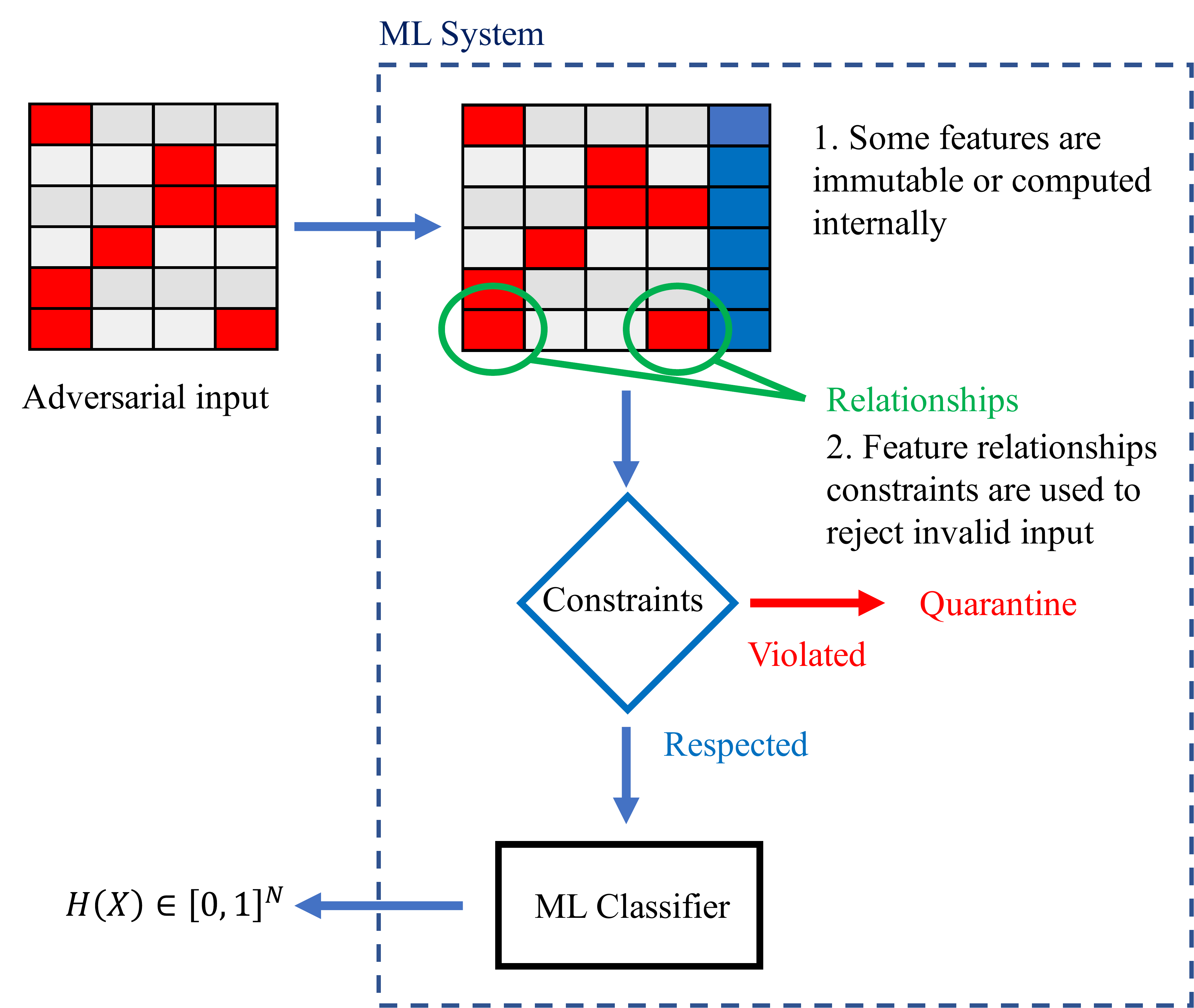}
    
    \caption{\small The main challenges for adversarial attacks in Tabular Machine Learning: When an adversary perturbs some features (red), it may not be aware of the new features that are computed internally and added (blue), or the relationships between features (green). If the monitoring system detects a constraint violation, the input is quarantined and a rejection (1) is returned.}
    \label{fig:challenges}
    \vspace{-1.5em}
\end{wrapfigure}

One significant challenge is that tabular data exhibit \emph{feature constraints}, which are complex relationships and interactions between features. Satisfying these feature constraints can be a non-convex or even nondifferentiable problem, making established evasion attack algorithms relying on gradient descent ineffective in generating valid adversarial examples (i.e., constraint-satisfying)~~\citep{ghamizi2020search}. Furthermore, attacks designed specifically for tabular data often disregard feature-type constraints ~\citep{ballet2019imperceptible} or, at best, consider categorical features without accounting for feature relationships ~\citep{wang2020attackability, xu2023probabilistic, bao2023towards}, and are evaluated on datasets that contain only such features. This limitation restricts their applicability to domains with heterogeneous feature types.

Moreover, tabular ML models often involve specific feature engineering, that is, "secret" and inaccessible to an attacker. For example, in credit scoring applications, the end user can alter a subset of model features, but the other features result from internal processing that adds domain knowledge before reaching the model~~\citep{ghamizi2020search}. This raises the need for new threat models that take into account these specificities. We summarize the unique specificities of tabular machine learning and the challenges they pose to an adversarial user in Figure \ref{fig:challenges}.

Thus, the machine learning research community currently lacks not only (1) an empirical understanding of the impact of architecture and robustification mechanisms on tabular data model architectures, but also (2) a reliable and high-quality benchmark to enable such investigations.
Such a benchmark for tabular adversarial attacks should feature deployable attacks and defenses that reflect as accurately as possible the robustness of models within a reasonable computational budget. A reliable benchmark should also consider recent advances in tabular deep learning architectures and data augmentation techniques, and tackle realistic attack scenarios and real-world use cases considering their domain constraints and realistic capabilities of an attacker.  

To address both gaps, we propose TabularBench, the first comprehensive benchmark of robustness of tabular deep learning classification models. We evaluated adversarial robustness using \emph{Constrained Adaptive Attack (CAA)}~\citep{simonetto2024constrained}, a combination of gradient-based and search-based attacks that has recently been shown to be the most effective against tabular models. We take advantage of our new benchmark and uncover unique findings on deep tabular learning architectures and defenses. We focus our study on defenses based on adversarial training (AT), and draw the following insights: %

\textbf{Test performance is misleading:} Given the same tasks, different architectures have similar ID performance but lead to very disparate robust performances. Even more, data augmentations that improve ID performance can hurt robust performance.

\textbf{Importance of domain constraints:} Disregarding domain constraints overstimates robustness and leads to selection of sub-optimal architectures and defenses when considering the domain constraints.

\textbf{Data augmentation effectiveness is task-specific.} There is no data augmentation that is optimal for both ID and robust performance across all tasks. Some simpler augmentations (like Cutmix) can outperform complex generative approaches.

\begin{figure}
    \centering
    \vspace{-1em}
    \includegraphics[clip, trim=1cm 1.5cm 1cm 2cm, width=\textwidth]{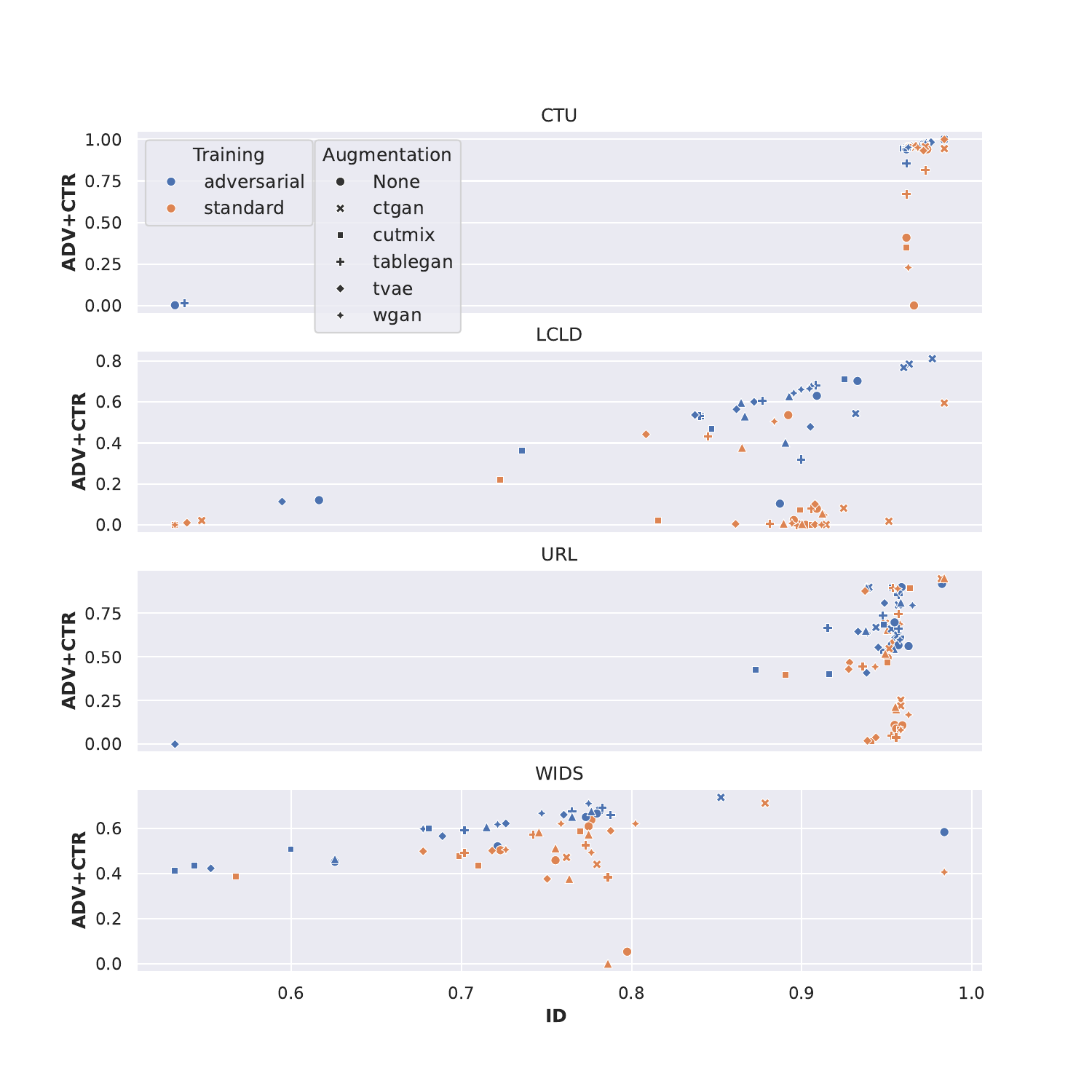}
    \caption{Summary of our main experiments; Y-axis: Robust Accuracy, X-axis ID accuracy}
    \label{fig:summary}
    \vspace{-1em}
\end{figure}

\paragraph{Contributions.}
To summarize, our work makes the following key contributions:
\begin{itemize}
    \item \textbf{Leaderboard} (\url{https://serval-uni-lu.github.io/tabularbench}): a website with a leaderboard based on \textit{more than 200} evaluations to track the progress and the current state of the art in adversarial robustness of tabular deep learning models for each critical setting. The goal is to clearly identify the most successful ideas in tabular architectures and robust training mechanisms to accelerate progress in the field. 
    \item \textbf{Dataset Zoo} %
    : a collection of real and synthetic datasets generated with and without domain-constraint satisfaction, over five critical tabular machine learning use cases.
    \item \textbf{Model Zoo} %
    : a collection of the most robust models that are easy to use for any downstream application. We pre-trained these models in particular on our five downstream tasks and we expect that this collection will promote the creation of more effective adversarial attacks by simplifying the evaluation process across a broad set of \textit{over 200} models. %
    \item \textbf{Analysis}: based on our trained models, we analyze how architectures, AT, and data augmentation mechanisms affect the robust performance of tabular deep learning models and provide insights on the best strategies per use case.
    
\end{itemize}

\section{Background}

Tabular data are one of the most common forms of data~\citep{Shwartz-Ziv2021}, especially in critical applications such as medical diagnosis~\citep{ulmer2020trust,somani2021deep} and financial applications~\citep{ghamizi2020search,cartella2021adversarial}. 

Traditional ML such as random forests and XGBoost often outperform DL on tabular data, primarily due to their robustness in handling feature heterogeneity and interdependence~\citep{Borisov_2022}. 

To bridge the gap, researchers have proposed various improvements, from regularization mechanisms (e.g., RLN~\citep{shavitt2018regularization}) to attention layers (TabNet~\citep{arik2021tabnet}). These innovations are catching up and even outperforming shallow models in some settings, demonstrating the competitiveness of DL for Tabular Data. 

The maturity of DL for ID tasks opens new perspectives for studying its performance in advanced settings, such as out-of-distribution (OOD) performance and adversarial robustness. One major work on OOD research is the Tableshift benchmark~\citep{gardner2023benchmarking}, an exhaustive evaluation of the OOD performance of a variety of DNN classifiers. There is, however, to the best of our knowledge, no similar work on adversarial robustness, while the use-cases when DL models are deployed for tabular data are among the most critical settings, and many are prone to malicious users.    

Our work is the first exhaustive benchmark for the critical property of adversarial robustness of DL models. Our work is timely and leverages CAA~\citep{simonetto2024constrained}, a novel attack previously demonstrated as the most effective and efficient tabular attack in the literature in multiple classification tasks under realistic constraints. CAA combines two attacks, CAPGD and MOEVA. CAPGD is an iterative gradient attack that maximizes the error and minimizes the features' constraint violations with regularization losses and projection mechanisms. MOEVA is a genetic algorithm attack that considers the three adversarial objectives: (1) classifier's error maximization, (2) perturbation minimization, and (3) constraint violations minimization, in its fitness function.

Although CAA was only evaluated against vanilla and simple madry AT, we have implemented advanced robustification mechanisms, inspired by proven techniques from top-performing research in the Robustbench computer vision benchmark Robustbench~\citep{croce2020robustbench}. Our work is the first implementation and evaluation of state-of-the-art defense mechanisms for tabular DL models.

\section{TabularBench: Adversarial Robustness Benchmark for Tabular Data}

In Appendix \ref{sec:app_eval_settings} we report the detailed evaluation settings such as metrics, attack parameters, and hardware. We focus below on the datasets, classifiers, and synthetic data generators.

\subsection{Tasks}
\label{sec:tasks}

We curated datasets meeting the following criteria:
(1) \textbf{open source:} the datasets must be publicly available with a clear definition of the features and preprocessing,
(2) \textbf{from real-world applications:} datasets that do not contain simulated data, (3) \textbf{binary classification:} datasets that support a meaningful binary classification task, and (4) \textbf{with feature relationships}: datasets that contain feature relationships and constraints, or they can be inferred directly from the definitions of features.   

After an extensive review of tabular datasets, only the following five datasets match our requirements.

The \textbf{CTU}~\citep{chernikova2022fence} includes legitimate and botnet traffic from CTU University. Its challenge lies in the extensive number of linear domain constraints, totaling 360.
\textbf{LCLD}~\citep{lcld} is a credit-scoring containing accepted and rejected credit requests. It has $28$ features and $9$ \emph{non-linear} constraints.
The most challenging dataset of our benchmark is the \textbf{Malware} dataset prepared by \citet{dyrmishi2022empirical}. 
The very large number of features ($24222$), most of which are involved in each constraint, make this dataset challenging to attack.
\textbf{URL}~\citep{hannousse2021towards} is a dataset comprising both legitimate and phishing URLs. Featuring only 14 linear domain constraints and 63 features, it represents the simplest case in our benchmark.
The \textbf{WiDS}~\citep{wids} includes medical data on the survival of patients admitted to the ICU, with only 31 linear domain constraints.

Our datasets include varying complexity in terms of number of features and constraints and diverse class imbalance intensity. We summarize the datasets and their relevant properties in Table \ref{tab:datasets} and provide more details in Appendix \ref{subsec:dataset_app} .  

\begin{table}[!ht]
    \small
    \centering    
    \caption{Properties of the use cases of our benchmark.}

    \begin{tabular}{c|c|c|c|c|c|c}
    \toprule
        Dataset & Domain & Output to flip & Total size & \# Features & \# Ctrs & Inbalance  \\
        \midrule
          CTU & Botnet detection & Malicious connections & 198 128 & 756 & 360 & 99.3/0.7\\ 
           LCLD & Credit scoring & Reject loan request & 1 220 092 & 28 & 9 &80/20\\ 
             Malware & Malware detection & Malicious software &  17 584  &24 222 & 7 & 45.5/54.5 \\ 
        URL & Phishing & Malicious URL & 11 430 & 63 & 14 & 50/50\\

        WIDS & ICU survival & Expected survival & 91 713 & 186 & 31 &91.4/8.6\\

        \bottomrule
    \end{tabular}
    \label{tab:datasets}
\end{table}

\subsection{Architectures}
We consider five state-of-the-art deep tabular architectures from the survey by \citet{borisov2021deep}:
\textbf{TabTransformer}~\citep{huang2020tabtransformer} and \textbf{TabNet}~\citep{arik2021tabnet}, are based on transformer architectures. \textbf{RLN}~\citep{shavitt2018regularization} uses a regularization coefficient to minimize a counterfactual loss, \textbf{STG}~\citep{icml2020_5085} improves feature selection using stochastic gates, while \textbf{VIME}~\citep{yoon2020vime} depends on self-supervised learning. We provide in Appendix \ref{sec:app_model_arch} the details of the architectures and the training hyperparameters.
These architectures are on par with XGBoost, the top shallow machine-learning model for our applications.

\subsection{Data Augmentation}
Our benchmark considers synthetic data augmentation using five state-of-the-art tabular data generators. These generators were pre-trained to learn the distribution of the training data. Then, we augmented each of our datasets 100-fold (for example, for URL dataset, we generated $1.143.000$ synthetic examples). Appendix \ref{sec:app_aug_arch} details the generator architectures and the training hyperparameters.

 \textbf{WGAN}~\citep{arjovsky2017wasserstein} is a typical generator-discriminator GAN model using Wasserstein loss. We follow the implementation of \citet{stoian2024how} and apply a MinMax transformation for continuous features and one-hot encoding for categorical to adapt this architecture for tabular data.
    
\textbf{TableGAN}~\citep{Park_2018} is an improvement over standard GAN generators for tabular data. It adds a classifier (trained to learn the labels and feature relationships) to the generator-discriminator setup to improve semantic accuracy. TableGAN uses MinMax transformation for features.

\textbf{CTGAN}~\citep{xu2019modeling} uses a conditional generator and training-by-sampling strategy in a generator-discriminator GAN architecture to model tabular data.

\textbf{TVAE}~\citep{xu2019modeling} is an adaptation of the Variational AutoEncoder architecture for tabular data. It uses the same data transformations as CTGAN and training with ELBO loss.

\textbf{GOGGLE}~\citep{liu2023goggle} is a graph-based model that learns relational and functional dependencies in data using graphs and a message passing DNN, generating variables based on their neighborhood.

\textbf{Cutmix}~\citep{yun2019cutmix} In computer vision, patches are cut and pasted among training images where the labels are also mixed proportionally. We adapted the approach to tabular ML and for each pair of rows of the same class, we randomly mix half of the features to generate a new sample.

For training, each batch of real examples is augmented with a same-size random synthetic batch (without replacement). However, the evaluation only runs on real examples.
In AT, we generate adversarials from half of the real examples randomly selected and half of the synthetic examples.

\subsection{TabularBench API}

To encourage the wide adoption of TabularBench as the go-to place for Tabular Machine Learning evaluation, we designed its API to be modular, extensible, and standardized. We split its architecture into three independent components. More details of each component are provided in Appendix \ref{sec:appendx_api}.

\textbf{A dataset Zoo} For each dataset in this study, we have collected, cleaned, and pre-processed the existing raw dataset. We implemented a novel \emph{Constraint Parser} where the user can write the relations in a natural human-readable format to describe the relationships between features. The processed datasets are loaded with a \emph{Dataset factory}, then the user gets their associated meta-data and pre-defined constraints. The datasets are automatically downloaded when not found.

\begin{python}
ds = dataset_factory.get_dataset("lcld_v2_iid")
metadata = ds.get_metadata(only_x=True)
constraints = ds.get_constraints()
\end{python}

\textbf{A model Zoo} Our API supports five architectures, and for each, six data augmentation techniques (as well as no data augmentation) and two training schemes (standard training and adversarial training). Hence, 70 pre-trained models for each of our five datasets are accessible. Below, we fine-tune with CAA AT and CTGAN augmentation a pre-trained Tabtransformer with Cutmix augmentation: 

\begin{python}
scaler = TabScaler(num_scaler="min_max", one_hot_encode=True)
scaler.fit(x, metadata["type"])
model = TabTransformer("regression", metadata, scaler=scaler,pretrained="LCLD_TabTr_Cutmix")
train_dataloader = CTGANDataLoader(dataset=ds, split="train", scaler=scaler, attack="caa")
model.fit(train_dataloader)
\end{python}

\textbf{A standarized benchmark} 
To generate our leaderboard, we offer a one-line command that loads a pre-trained model from the zoo, and reports the clean and robust accuracy of the model following our benchmark's setting (taking into consideration constraint satisfaction and L2 minimization):
\begin{python}
clean_acc, robust_acc = benchmark(dataset='LCLD', model="TabTr_Cutmix", distance='L2', constraints=True)
\end{python}

\section{Empirical Findings}

In the main paper, we provide multiple figures to visualize the main insights. We only report scenarios where data augmentation and adversarial training do not lead to performance collapse. We report in Appendix \ref{sec:detailed-results} all the results and investigate the collapsed scenarios.

\begin{figure}[!ht]
    \centering
    \includegraphics[clip, trim=8.5cm 5.5cm 8.5cm 5.5cm, width=\textwidth]{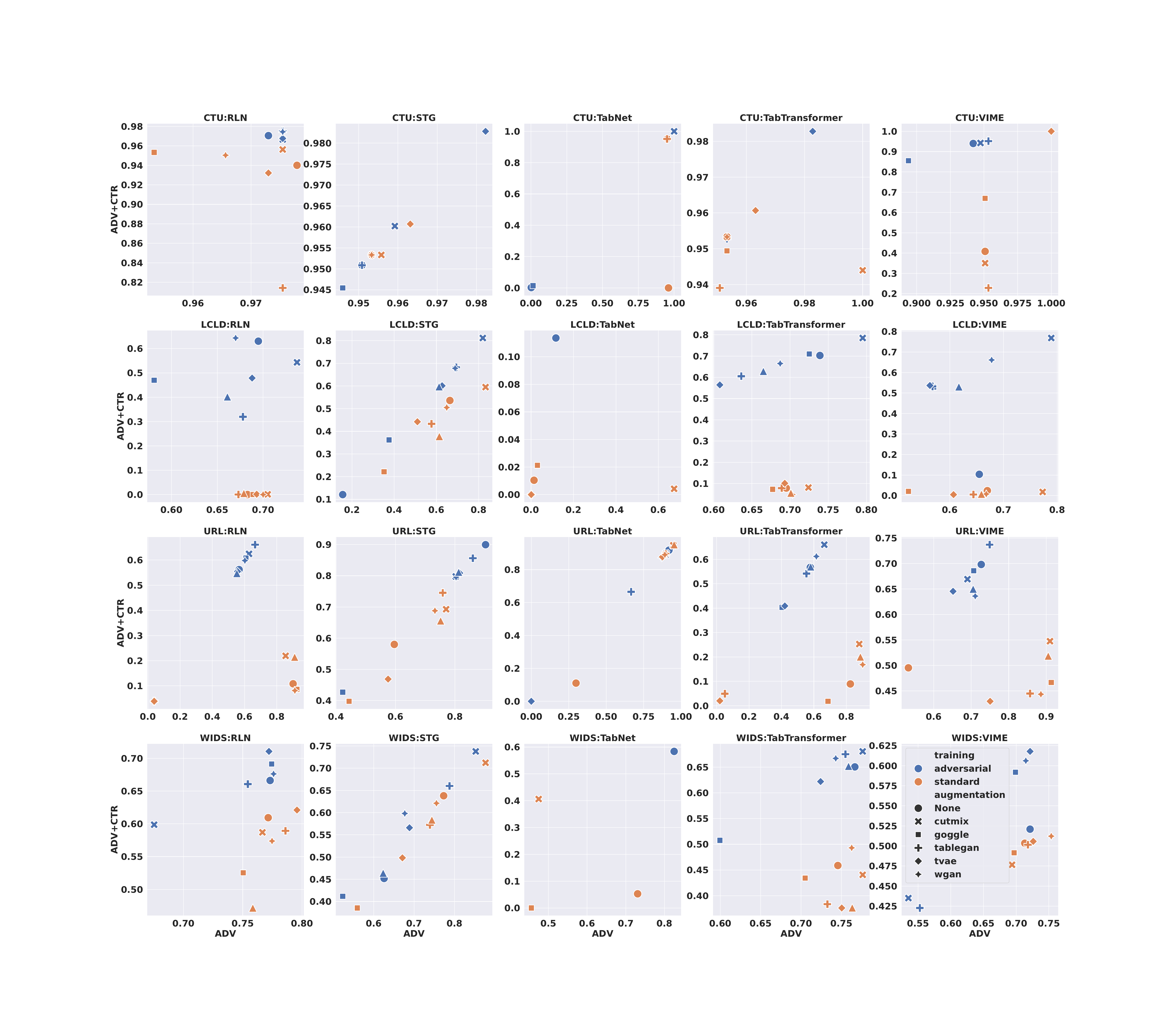}
    \caption{Robust performance while considering domain constraints (ADV+CTR: Y-axis) and without (ADV: X-axis) on all our use cases confirms the relevance of studying constrained-aware attacks.}
    \label{fig:arch_perf}
\end{figure}

\subsection{Without Data Augmentations}

We report the ID and robust accuracies of our architectures prior to data increase in Table \ref{tab:defense}.

\begin{table}[!ht]
\centering
\caption{Clean and robust performances across all architectures in the form XX/YY. XX is the accuracy with standard training, and YY is the accuracy with adversarial training.}
\label{tab:defense}
\small
\begin{tabular}{ll|rrrrr}
\toprule
Dataset & Accuracy & TabTr. & RLN & VIME & STG & TabNet \\
\midrule
\multirow[c]{2}{*}{CTU} & ID & $95.3/95.3$ & $97.8/97.3$ & $95.1/95.1$ & $95.3/95.1$ & $96.0/0.2$ \\
 & Robust & $95.3/95.3$ & $94.1/97.1$ & $40.8/94.0$ & $95.3/95.1$ & $0.0/0.2$ \\
\cline{1-7}

\multirow[c]{2}{*}{LCLD} & ID & $69.5/73.9$ & $68.3/69.5$ & $67.0/65.5$ & $66.4/15.6$ & $67.4/0.0$ \\
 & Robust & $7.9/70.3$ & $0.0/63.0$ & $2.4/10.4$ & $53.6/12.1$ & $0.4/0.0$ \\
\cline{1-7}
\multirow[c]{2}{*}{MALWARE} & ID & $95.0/95.0$ & $95.0/96.0$& $95.0/92.0$ & $93.0/93.0$ & $99.0/99.0$ \\
 & Robust & $94.0/95.0$ & $94.0/96.0$ & $95.0/92.0$ & $93.0/93.0$ & $97.0/99.0$  \\
\cline{1-7}
\multirow[c]{2}{*}{URL} & ID & $93.6/93.9$ & $94.4/95.2$ & $92.5/93.4$ & $93.3/94.3$ & $93.4/99.5$ \\
 & Robust & $8.9/56.7$ & $10.8/56.2$ & $49.5/69.8$ & $58.0/90.0$ & $11.0/91.8$ \\
\cline{1-7}
\multirow[c]{2}{*}{WIDS} & ID & $75.5/77.3$ & $77.5/78.0$ & $72.3/72.1$ & $77.7/62.6$ & $79.8/98.4$ \\
 & Robust & $45.9/65.1$ & $60.9/66.6$ & $50.3/52.1$ & $50.3/45.2$ & $5.3/58.4$ \\

\bottomrule

\end{tabular}
\end{table}

\textbf{All models on malware dataset are robust without data augmentation.} AT improves adversarial accuracy for all the cases, but AT alone is not sufficient to completely robustify the models on URL and WIDS datasets.
All malware classification models are completely robust with and without adversarial training; 
hence, we will restrict the study of improved defenses with augmentation in the following sections to the remaining datasets.

\subsection{Impact of Data Augmentations}

\textbf{With data augmentation alone, ID and robust performances are not aligned.} In Figure \ref{fig:summary} we study the impact of data augmentation on ID and robust performance, both in standard and adversarial training. 
With standard training, ID performance is misleading in CTU and URL datasets. Although all models exhibit similar ID performance, some of the augmentations lead to robust models, while others decrease it. CTGAN data augmentation is the best data augmentation for ID performance in all use cases, both with standard and adversarial training.

\subsection{Impact of Adversarial Training}

\textbf{With data augmentation and AT, ID and robust performances are correlated.} Although there is no trend of relationship between ID performance and robust performance in standard training, our study shows that robustness and ID performance are correlated after adversarial training. For example, the Pearson correlation between ID and robust performance increases from $0.15$ to $0.76$ for LCLD. All correlation values are in Appendix \ref{subsec:appendix-correlations}.

Overall, all architectures can benefit from at least one data augmentation technique with adversarial training; however, standard training with data augmentation can outperform adversarial training without data augmentation (for e.g., on URL dataset using GOGGLE or CTGAN augmentations).

\subsection{Impact of Architecture}

In Figure \ref{fig:arch_perf} we study the robustness of each architecture with different defense mechanisms. We report both the robustness against unconstrained attacks (attacks unaware of domain knowledge) and attacks optimized to preserve the feature relationships and constraints.

\textbf{Evaluation with unconstrained attack is misleading.} Under standard training (orange scatters in Fig. \ref{fig:arch_perf}), there is no relation between robustness to unconstrained attacks and the robustness when domain constraints are enforced. 
There is, however, a linear relationship under adversarial training with data augmentation only for STG, Tabstransformer and VIME architectures. 
These results show that nonconstrained attacks are not sufficient to reliably assess the robustness of deep tabular models. 
Detailed correlation values are in the Appendix \ref{subsec:appendix-correlations}.

\textbf{No data augmentation consistently outperforms the baselines with AT.}
Among the 20 scenarios in Fig. \ref{fig:arch_perf}, the original models achieve better constrained robustness than augmented models with adversarial training only for 4 scenarios: TabNet architecture on URL, LCLD and WIDS, and STG architecture on URL datasets. No data-augmentation technique consistently outperforms the others across all architectures. Cutmix, the simplest data augmentation, is often the best (in 7/20 scenarios).    

\subsection{Impact of Attack Budgets}

\begin{figure}[t]
\centering
\begin{subfigure}{0.29\textwidth}
    \includegraphics[clip, trim=0.6cm 1.4cm 0.6cm 0.2cm,width=\textwidth]{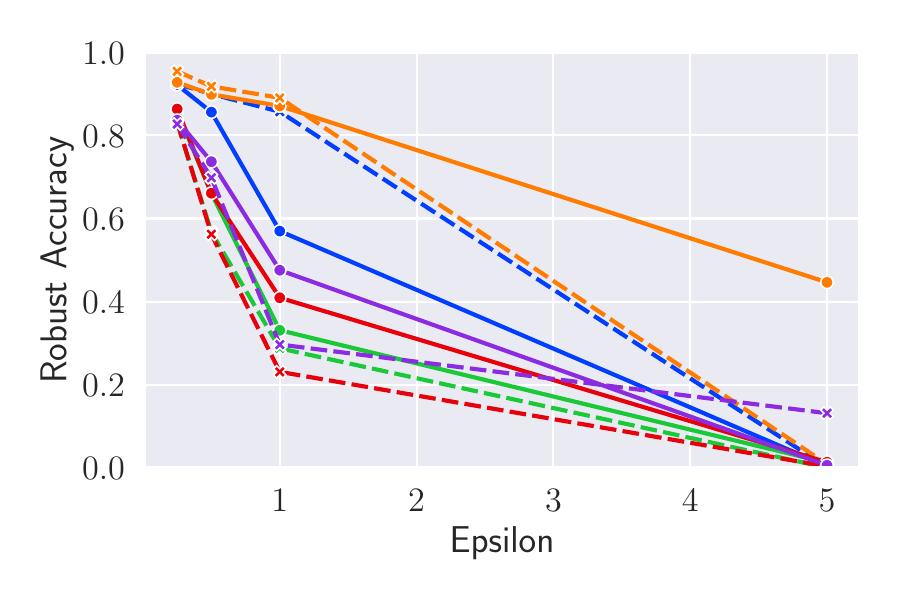}
    \caption{Maximum $\epsilon$ perturbation}
    \label{fig:caa-url-eps}
\end{subfigure}
\hfill
\begin{subfigure}{0.29\textwidth}
    \includegraphics[clip, trim=0.6cm 1.4cm 0.6cm 0.2cm,width=\textwidth]{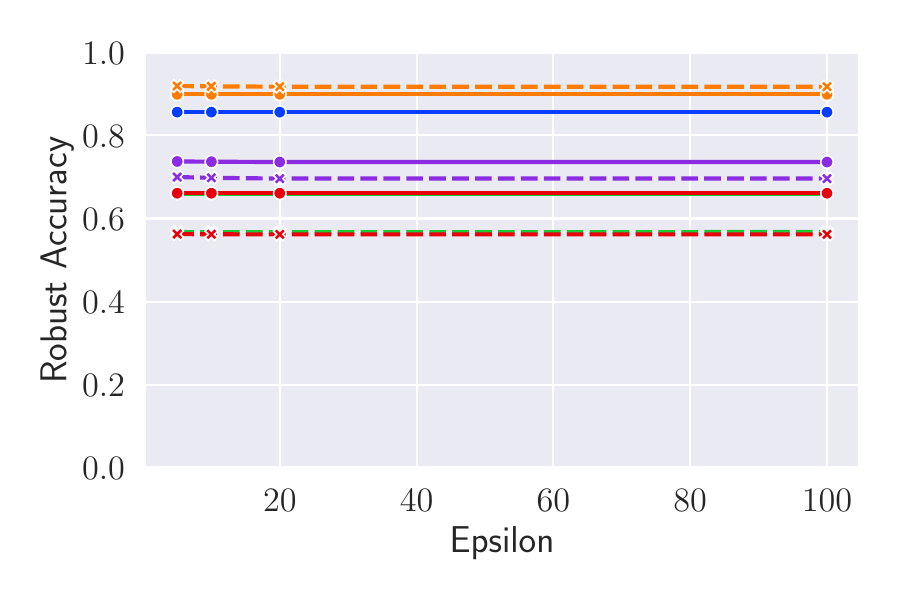}
    \caption{Gradient attack iterations}
    \label{fig:caa-url-iter-gradient}
\end{subfigure}
\hfill
\begin{subfigure}{0.29\textwidth}
    \includegraphics[clip, trim=0.6cm 1.4cm 0.6cm 0.0cm,width=\textwidth]{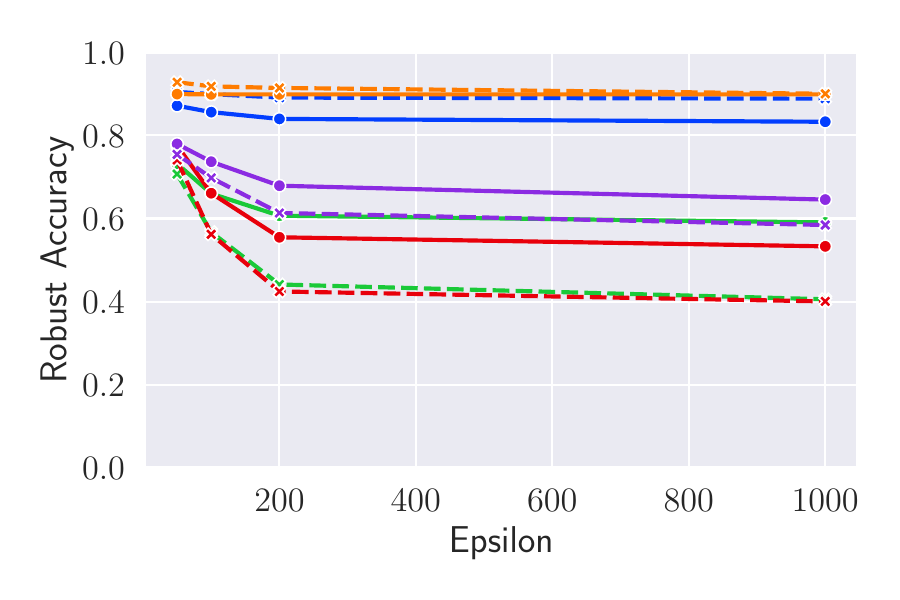}
    \caption{Search attack iterations}
    \label{fig:caa-url-iter-search}
\end{subfigure}
\begin{subfigure}{0.10\textwidth}
    \includegraphics[clip, width=\textwidth]{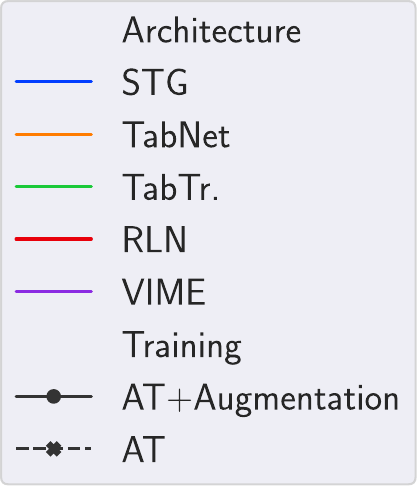}
    \caption*{}
    \label{fig:caa-url-legend}
\end{subfigure}
\caption{Impact of attack budget on the robust accuracy for URL dataset.}
\label{fig:caa-url-budget}
\end{figure}

We evaluated each robustified model against variants of the CAA attack, varying the $L_2$ distance of the perturbation $\epsilon$ from $0.5$ to $\{0.25,1,5\}$, the gradient iterations from $10$ to $\{5,20,100\}$, and the search iterations from $100$ to $\{50,200,1000\}$. We report per architecture for each dataset the most robust model with AT and augmentation, and the robust model with AT only. We present in Fig. \ref{fig:caa-url-budget} the results for the URL dataset and refer to Appendix \ref{subsec:appendix-budgets} for the other use cases.

\textbf{AT+Augmentations models remain robust even under stronger attacks.} Our results show that the best defenses with AT+Augmentations (continuous lines) remain robust against increased gradient and search iteration budgets and remain more robust than AT alone (dashed lines) for VIME, RLN, and Tabtransformer architectures. Against an increase in perturbation size $\epsilon$, AT+Augmentations is more robust than AT alone for TabNet, TabTransformer, VIME, and RLN architectures. In particular, for $\epsilon=5$, the robust accuracy of TabNet architectures remains above 40\% with AT+Augmentations while the robust accuracy with AT alone drops to 0\%. 

\section{Limitations}
\label{sec:limitations}
While our benchmark is the first to tackle adversarial robustness in tabular deep learning models, it does not cover all the directions of the field and focuses on domain constraints and defense mechanisms. Some of the orthogonal work is not addressed:

\textbf{Generalization to other distances:} We restricted our study to the $L_2$ distance to measure imperceptibility. Imperceptibility varies by domain, and several methods have been proposed to measure it \citep{ballet2019imperceptible,kireev2022adversarial, dyrmishi2022empirical}. These methods have not been evaluated against human judgment or compared with one another, so there is no clear motivation to use one or another. In our research, we chose to use the well-established $L_2$ norm (following \citet{dyrmishi2022empirical}). Our algorithms and benchmarks support other distances and definitions of imperceptibility. We provide in Appendix \ref{subsec:appendix-distance} an introduction to how our benchmark generalizes to other distances.

\textbf{Generalization to non-binary classification:} We restricted our study to binary tabular classification as it is the only case where we identified public datasets with domain constraints. The attacks used in our benchmark natively support multi-class classification. Our live leaderboard welcomes new datasets and will be updated if relevant datasets are designed by the community.

\textbf{Generalization to other types of defenses:} We only considered defenses based on data augmentation with adversarial training. Adversarial training based defenses are recognized as the only reliable defenses against evasion attack~\citep{tramer2020adaptive,carlini2023llm}. All other defenses are proven ineffective when the attacker is aware of them and performs adaptive attacks.

\section{Broader Impact}
\label{sec:impact}
Our work proposes the first benchmark of robustness of constrained tabular deep learning against evasion attacks.
We focus on designing new defense mechanisms, inspired by effective approaches in computer vision (by combining data augmentation and adversarial training). Hence, we expect that our research will significantly contribute to the enhancement of defenses and will lead to even more resilient models, which may balance the potential harms research on adversarial attacks can have.

\section*{Conclusion}

In this work, we introduce TabularBench, the first benchmark of adversarial robustness of tabular deep learning models against constrained evasion attacks.
We leverage Constrained Adaptive Attack (CAA), the best constrained tabular attack, to benchmark state-of-the-art architectures and defenses. 

We provide a Python API to access the datasets, along with implementations of multiple tabular deep learning architectures, and provide all our pretrained robust models directly through the API. 

We conducted an empirical study that constitutes the first large-scale study of tabular data model robustness against evasion attacks. Our study covers five real-world use cases, five architectures, and six data augmentation mechanisms totaling more than 200 models. Our study identifies the best augmentation mechanisms for IID performance (CTGAN) and robust performance (Cutmix), and provides actionable insights on the selection of architectures and robustification mechanisms.

We are confident that our benchmark will accelerate the research of adversarial defenses for tabular ML and welcome all contributions to improve and extend our benchmark with new realistic use cases (multiclass), models, and defenses.

\clearpage

\bibliographystyle{ACM-Reference-Format}
\bibliography{bib/adv, bib/blackbox, bib/tabsurvey, bib/realistic, bib/benchmarks}
\newpage

\clearpage

\appendix

\section{Experimental protocol}
\label{sec:experimental-protocol}

\subsection{Datasets}
\label{subsec:dataset_app}

\begin{table*}[t]
  \centering
  \caption{The datasets evaluated in the empirical study, with the class imbalance of each dataset.}
  \label{tab:data_extended}
  \begin{tabular}{l|llll}
    \toprule
    Dataset & \multicolumn{4}{c}{Properties} \\
     & Task & Size & \# Features & Balance (\%) \\
    \midrule
    LCLD~\cite{lcld} &  Credit Scoring &  1 220 092 & 28 & 80/20 \\
    CTU-13~\cite{chernikova2022fence}&     Botnet Detection    &   198 128 & 756 & 99.3/0.7  \\
    URL~\cite{hannousse2021towards} & Phishing URL detection    & 11 430 & 63  & 50/50 \\
    WIDS~\cite{wids} & ICU patient survival & 91 713 & 186& 91.4/8.6 \\
    \bottomrule
  \end{tabular}
\end{table*}

Our dataset design followed the same protocol as Simonetto et al.\cite{simonetto2021unified}.
We present in Table \ref{tab:data_extended} the attributes of our datasets and the test performance achieved by each of the architectures.

\paragraph{Credit Scoring - LCLD} (license: CC0: Public Domain) We develop a dataset derived from the publicly accessible Lending Club Loan Data\\footnote{https://www.kaggle.com/wordsforthewise/lending-club}. This dataset includes 151 features, with each entry representing a loan approved by the Lending Club. However, some of these approved loans are not repaid and are instead charged off. Our objective is to predict, at the time of the request, whether the borrower will repay the loan or if it will be charged off. This dataset has been analyzed by various practitioners on Kaggle. Nevertheless, the original dataset only contains raw data, and to the best of our knowledge, there is no commonly used feature-engineered version. Specifically, caution is needed when reusing feature-engineered versions, as many proposed versions exhibit data leakage in the training set, making the prediction trivial. Therefore, we propose our own feature engineering. The original dataset contains 151 features. We exclude examples where the feature ``loan status'' is neither ``Fully paid'' nor ``Charged Off,'' as these are the only definitive statuses of a loan; other values indicate an uncertain outcome. For our binary classifier, a ``Fully paid'' loan is represented as 0, and a ``Charged Off'' loan is represented as 1. We begin by removing all features that are missing in more than 30\% of the examples in the training set. Additionally, we remove all features that are not available at the time of the loan request to avoid bias. We impute features that are redundant (e.g., grade and sub-grade) or too detailed (e.g., address) to be useful for classification. Finally, we apply one-hot encoding to categorical features. We end up with 47 input features and one target feature. We split the dataset using random sampling stratified by the target class, resulting in a training set of 915K examples and a testing set of 305K examples. Both sets are unbalanced, with only 20\% of loans being charged off (class 1). We trained a neural network to classify accepted and rejected loans, consisting of 3 fully connected hidden layers with 64, 32, and 16 neurons, respectively. For each feature in this dataset, we define boundary constraints based on the extreme values observed in the training set. We consider the 19 features under the control of the Lending Club as immutable. We identify 10 relationship constraints (3 linear and 7 non-linear).

\paragraph{URL Phishing - ISCX-URL2016} (license CC BY 4.0) Phishing attacks are commonly employed to perpetrate cyber fraud or identity theft. These attacks typically involve a URL that mimics a legitimate one (e.g., a user's preferred e-commerce site) but directs the user to a fraudulent website that solicits personal or banking information. \cite{hannousse2021towards} extracted features from both legitimate and fraudulent URLs, as well as external service-based features, to develop a classifier capable of distinguishing between fraudulent and legitimate URLs. The features extracted from the URL include the number of special substrings such as ``www'', ``\&'', ``,'', ``\$'', ``and'', the length of the URL, the port, the presence of a brand in the domain, subdomain, or path, and the inclusion of ``http'' or ``https''. External service-based features include the Google index, page rank, and the domain's presence in DNS records. The full list of features is available in the reproduction package. \cite{hannousse2021towards} provide a dataset containing 5715 legitimate and 5715 malicious URLs. We use 75\% of the dataset for training and validation, and the remaining 25\% for testing and adversarial generation. We extract a set of 14 relational constraints between the URL features. Among these, 7 are linear constraints (e.g., the length of the hostname is less than or equal to the length of the URL) and 7 are Boolean constraints of the form $if\ a > 0 \ then\ b > 0$ (e.g., if the number of ``http'' $>$ 0, then the number of slashes ``/'' $>$ 0).
\paragraph{Botnet attacks - CTU-13} (license CC BY NC SA 4.0) This is a feature-engineered version of CTU-13 proposed by \cite{chernikova2019fence}. It includes a combination of legitimate and botnet traffic flows from the CTU University campus. Chernikova et al. aggregated raw network data related to packets, duration, and bytes for each port from a list of commonly used ports. The dataset consists of 143K training examples and 55K testing examples, with 0.74\% of examples labeled as botnet traffic (traffic generated by a botnet). The data contains 756 features, including 432 mutable features. We identified two types of constraints that define what constitutes feasible traffic data. The first type pertains to the number of connections and ensures that an attacker cannot reduce it. The second type involves inherent constraints in network communications (e.g., the maximum packet size for TCP/UDP ports is 1500 bytes). In total, we identified 360 constraints.
\paragraph{WiDS} (license: PhysioNet Restricted Health Data License 1.5.0 \footnote{https://physionet.org/content/widsdatathon2020/view-license/1.0.0/}) \cite{wids} dataset contains medical data on the survival of patients admitted to the ICU. The objective is to predict whether a patient will survive or die based on biological features (e.g., for triage). This highly unbalanced dataset has 30 linear relational constraints.

\paragraph{Malware} (licence MIT) contains 24222 features extracted from a collection of benign and malware Portable Executable (PE) files \cite{dyrmishi2022empirical}. The features include the DLL imports, the API imports, PE sections, and statistic features such as the proportion of each possible byte value. The dataset contains 17,584 samples. 
The number of total features and the number of features involved in each constraint make this dataset challenging to attack. The objective of the classifier is to distinguish between malware and benign software.

\subsection{Model architectures}
\label{sec:app_model_arch}
\begin{table*}
\centering
  \caption{The three model architectures of our study.}
  \label{tab:models}
  \small
  \begin{tabular}{lll}
  
    \toprule
    Family & Model & Hyperparameters\\
    \midrule

    Transformer & TabTransformer  & \begin{tabular}[c]{@{}l@{}}$hidden\_dim$, $n\_layers$,\\ $learning\_rate$, $norm$, $\theta$\end{tabular}  \\
    Transformer & TabNet  & \begin{tabular}[c]{@{}l@{}}$n\_d$, $n\_steps$,\\ $\gamma$, $cat\_emb\_dim$, $n\_independent$, \\ $n\_shared$, $momentum$, $mask\_type$ \end{tabular}  \\
    Regularization & RLN & \begin{tabular}[c]{@{}l@{}}$hidden\_dim$, $depth$, \\ $heads$, $weight\_decay$, \\ $learning\_rate$, $dropout$\end{tabular} \\
    Regularization & STG & \begin{tabular}[c]{@{}l@{}} $hidden\_dims$, $learning\_rate$, $lam$\end{tabular} \\
        Encoding & VIME  &  $p_m$, $\alpha$, $K$, $\beta$\\

  \bottomrule
\end{tabular}
\end{table*}

Table \ref{tab:models} provides an overview of the family, model architecture, and hyperparameters adjusted during the training of our models.\paragraph{TabTransformer} is a transformer-based model~\cite{huang2020tabtransformer}. It employs self-attention to convert categorical features into an interpretable contextual embedding, which the paper asserts enhances the model's robustness to noisy inputs.\paragraph{TabNet} is another transformer-based model~\cite{arik2021tabnet}. It utilizes multiple sub-networks in sequence. At each decision step, it applies sequential attention to select which features to consider. TabNet combines the outputs of each step to make the final decision.\paragraph{RLN} or Regularization Learning Networks~\cite{shavitt2018regularization} employs an efficient hyperparameter tuning method to minimize counterfactual loss. The authors train a regularization coefficient for the neural network weights to reduce sensitivity and create very sparse networks.\paragraph{STG} or Stochastic Gates~\cite{icml2020_5085} uses stochastic gates for feature selection in neural network estimation tasks. The technique is based on a probabilistic relaxation of the $l_0$ norm of features or the count of selected features.\paragraph{VIME} or Value Imputation for Mask Estimation~\cite{yoon2020vime} employs self-supervised and semi-supervised learning through deep encoders and predictors.

\subsection{Evaluation settings}
\label{sec:app_eval_settings}

\paragraph{Metrics}
\label{sec:app_metrics}

The models are fine-tuned to maximize cross-validation AUC. This metric is threshold-independent and is not affected by the class unbalance of our dataset.

We only attack clean examples that are not already misclassified by the model and from the critical class, that is respectively for each aforementioned dataset the class of phishing URLs, rejected loans, malwares, botnets, and not surviving patients.
Because we consider a single class, the only relevant metric is robust accuracy on constrained examples.
Unsuccessful adversarial examples count as correctly classified when measuring robust accuracy.

We only consider examples that respect domain constraints to compute robust accuracy. If an attack generates invalid examples, they are defacto considered unsuccessful and are reverted to their original example (correctly classified). 

We report in the Appendix \ref{tab:detailed_clean_agumentation} all the remaining performance metrics, including the recall, the precision, and the Mattheu Correlation Coefficient (MCC).

\paragraph{Attacks parameters}
\label{sec:app_atk_params}
CAA applies CAPGD and MOEVA with the following parameters.

CAPGD uses $N_{iter} = 10$ iterations. The step reduction schedule for CPGD uses $M = 7$. In CAPGD, checkpoints are set as $w_j = \lceil p_j \times N_{iter} \rceil \leq N_{iter}$, with $p_j \in [0,1]$ defined as $p_0 = 0$, $p_1 = 0.22$, and \[p_{j+1} = p_j + \max\\{p_j - p_{j-1} - 0.03, 0.06\\}.\] 

The influence of the previous update on the current update is set to $\alpha = 0.75$, and $\rho = 0.75$ for step halving. MOEVA runs for $n_{gen} = 100$ iterations, generating $n_{off} = 100$ offspring per iteration. Among the offspring, $n_{pop} = 200$ survive and are used for mating in the subsequent iteration. 

\paragraph{Hardware}
\label{sec:hardware}

Our experiments are conducted on an HPC cluster node equipped with 32 cores and 64GB of RAM allocated for our use. Each node is composed of 2 AMD Epyc ROME 7H12 processors running at 2.6 GHz, providing a total of 128 cores and 256 GB of RAM.

\subsection{Generator architectures}
\label{sec:app_aug_arch}

In our experimental study, we use the same {five} generative models as \citet{stoian2024how}:

\begin{itemize}\item \textbf{WGAN}~\citep{arjovsky2017wasserstein} is a GAN model trained with Wasserstein loss within a standard generator-discriminator GAN framework. In our implementation, WGAN utilizes a MinMax transformer for continuous features and one-hot encoding for categorical features. It is not specifically designed for tabular data.\item \textbf{TableGAN}~\citep{Park_2018} is one of the pioneering GAN-based methods for generating tabular data. Besides the conventional generator and discriminator setup in GANs, the authors introduced a classifier trained to understand the relationship between labels and other features. This classifier ensures a higher number of semantically correct generated records. TableGAN applies a MinMax transformer to the features.\item \textbf{CTGAN}~\citep{Xu2019CTGAN} employs a conditional generator and a training-by-sampling strategy within a generator-discriminator GAN framework to model tabular data. The conditional generator produces synthetic rows conditioned on one of the discrete columns. The training-by-sampling method ensures that data are sampled according to the log-frequency of each category, aiding in better modeling of imbalanced categorical columns. CTGAN uses one-hot encoding for discrete features and a mode-based normalization for continuous features. A variational Gaussian mixture model~\citep{camino2018_vgm} is used to estimate the number of modes and fit a Gaussian mixture. For each continuous value, a mode is sampled based on probability densities, and its mean and standard deviation are used for normalization.\item {\textbf{TVAE}~\citep{Xu2019CTGAN} was introduced as a variant of the standard Variational AutoEncoder to handle tabular data. It employs the same data transformations as CTGAN and trains the encoder-decoder architecture using evidence lower-bound (ELBO) loss.}\item {\textbf{GOGGLE}~\citep{liu2023goggle} is a graph-based method for learning the relational structure of data as well as functional relationships (dependencies between features). The relational structure is learned by constructing a graph where nodes represent variables and edges indicate dependencies between them. Functional dependencies are learned through a message-passing neural network (MPNN). The generative model generates each variable considering its surrounding neighborhood.}\end{itemize}

The hyperameters for training these models is based on \citet{stoian2024how} as well:

\paragraph{For GOGGLE,} we employed the same optimizer and learning rate configuration as described in \cite{liu2023goggle}. Specifically, ADAM was used with five different learning rates: $\{ \num{1e-3}, \num{5e-3}, \num{1e-2} \}$. 

\paragraph{For TVAE,} ADAM was utilized with five different learning rates: $\{ \num{5e-6}, \num{1e-5}, \num{1e-4}, \num{2e-4}, \num{1e-3} \}$. 

For the other DGM models, three different optimizers were tested: ADAM, RMSPROP, and SGD, each with distinct sets of learning rates. 

\paragraph{For WGAN,} the learning rates were $\{ \num{1e-4}, \num{1e-3} \}$, $\{ \num{5e-5}, \num{1e-4}, \num{1e-3} \}$, and $\{ \num{1e-4}, \num{1e-3} \}$, respectively. 

\paragraph{For TableGAN,} the learning rates were $\{ \num{5e-5}, \num{1e-4}, \num{2e-4}, \num{1e-3} \}$, $\{ \num{1e-4}, \num{2e-4}, \num{1e-3} \}$, and $\{ \num{1e-4}, \num{1e-3} \}$, respectively. 

\paragraph{For CTGAN,} the learning rates were $\{ \num{5e-5}, \num{1e-4}, \num{2e-4} \}$, $\{ \num{1e-4}, \num{2e-4}, \num{1e-3} \}$, and $\{ \num{1e-4}, \num{1e-3} \}$, respectively. 

For each optimizer-learning rate combination, three different batch sizes were tested, depending on the DGM model: $\{64, 128, 256\}$ for WGAN, $\{128, 256, 512\}$ for TableGAN, $\{70, 280, 500\}$ for CTGAN and TVAE, and $\{64, 128\}$ for GOGGLE. The batch sizes for CTGAN are multiples of 10 to accommodate the recommended PAC value of 10 as suggested in \cite{lin2018pacgan}, among other values.

\subsection{Reproduction package and availability}
\label{sec:reproduction}

The source code, datasets, and pre-trained models required to replicate the experiments in this paper are publicly accessible under the MIT license on the repository \url{https://github.com/serval-uni-lu/tabularbench}.

\section{Detailed results}
\label{sec:detailed-results}

\subsection{Baseline models performances}

We compare in \ref{tab:app-iid-perf} the ID performance of XGBoost and our deep learning models under standard training. We confirm that DL models are on par with the performances achieved by shallow models.

\begin{table}[!h]
    \centering
    
    \caption{AUC In-distribution performance of models }
    \begin{tabular}{c|ccccc}
    \toprule
        Dataset &  CTU & LCLD & MALWARE & URL & WIDS  \\
        \midrule
        RLN &  0.991 & 0.719 & 0.993 & 0.984 & 0.869  \\
        STG &  0.988 & 0.709 & 0.991 & 0.973 & 0.866  \\
        TabNet &  0.996 & 0.722 & 0.994 & 0.986 & 0.870  \\
        TabTr &  0.979 & 0.717 & 0.994 & 0.981 & 0.874  \\
        VIME &  0.987 & 0.714 & 0.989 & 0.974 & 0.865  \\
        \midrule
        XGBoost &  0.994 & 0.723 & 0.997 & 0.993 & 0.887  \\
        
    \end{tabular}
    \label{tab:app-iid-perf}
\end{table}

\subsection{Data augmentation detailed results}

\paragraph{Clean performance after data augmentation}

We report in Table \ref{tab:detailed_clean_agumentation} the clean performances of our models under all the trianing scenarios. Notably, few training combinations lead to a collapse of performance ($MCC = 0$). It is the case on CTU dataset for all data augmentations with adversarial training, and CTGAN, Cutmix, and TVAE with standard training.

\begin{longtable}{llrrrrrll}[b]

\label{tab:detailed_clean_agumentation} \\
\caption{Detailed results of clean performance for our augmented models}\\    %

Dataset & Arch & AUC & Accuracy & Precision & Recall & Mcc & Training & Augment \\
\midrule
URL & TabTr & 0.981 & 0.940 & 0.943 & 0.937 & 0.880 & Standard & None \\
URL & TabTr & 0.974 & 0.931 & 0.923 & 0.941 & 0.862 & Adversarial & None \\
URL & TabTr & 0.976 & 0.933 & 0.927 & 0.941 & 0.866 & Standard & ctgan \\
URL & TabTr & 0.963 & 0.916 & 0.903 & 0.932 & 0.832 & Adversarial & ctgan \\
URL & TabTr & 0.968 & 0.930 & 0.954 & 0.905 & 0.862 & Standard & cutmix \\
URL & TabTr & 0.956 & 0.900 & 0.937 & 0.857 & 0.803 & Adversarial & cutmix \\
URL & TabTr & 0.974 & 0.931 & 0.932 & 0.930 & 0.862 & Standard & goggle \\
URL & TabTr & 0.964 & 0.915 & 0.913 & 0.918 & 0.830 & Adversarial & goggle \\
URL & TabTr & 0.980 & 0.934 & 0.934 & 0.934 & 0.869 & Standard & wgan \\
URL & TabTr & 0.970 & 0.921 & 0.916 & 0.927 & 0.843 & Adversarial & wgan \\
URL & TabTr & 0.975 & 0.928 & 0.955 & 0.899 & 0.858 & Standard & tablegan \\
URL & TabTr & 0.967 & 0.919 & 0.935 & 0.900 & 0.839 & Adversarial & tablegan \\
URL & TabTr & 0.978 & 0.937 & 0.925 & 0.950 & 0.873 & Standard & tvae \\
URL & TabTr & 0.969 & 0.925 & 0.917 & 0.934 & 0.850 & Adversarial & tvae \\
URL & STG & 0.973 & 0.920 & 0.908 & 0.934 & 0.839 & Standard & None \\
URL & STG & 0.949 & 0.862 & 0.812 & 0.943 & 0.734 & Adversarial & None \\
URL & STG & 0.967 & 0.910 & 0.898 & 0.925 & 0.820 & Standard & ctgan \\
URL & STG & 0.959 & 0.895 & 0.863 & 0.940 & 0.794 & Adversarial & ctgan \\
URL & STG & 0.960 & 0.867 & 0.924 & 0.800 & 0.741 & Standard & cutmix \\
URL & STG & 0.954 & 0.842 & 0.909 & 0.760 & 0.694 & Adversarial & cutmix \\
URL & STG & 0.962 & 0.903 & 0.876 & 0.940 & 0.809 & Standard & goggle \\
URL & STG & 0.954 & 0.882 & 0.842 & 0.941 & 0.770 & Adversarial & goggle \\
URL & STG & 0.970 & 0.913 & 0.903 & 0.926 & 0.826 & Standard & wgan \\
URL & STG & 0.963 & 0.896 & 0.862 & 0.943 & 0.796 & Adversarial & wgan \\
URL & STG & 0.968 & 0.908 & 0.933 & 0.878 & 0.817 & Standard & tablegan \\
URL & STG & 0.956 & 0.888 & 0.862 & 0.923 & 0.777 & Adversarial & tablegan \\
URL & STG & 0.969 & 0.913 & 0.892 & 0.940 & 0.827 & Standard & tvae \\
URL & STG & 0.961 & 0.889 & 0.843 & 0.956 & 0.786 & Adversarial & tvae \\
URL & TabNet & 0.986 & 0.946 & 0.954 & 0.937 & 0.892 & Standard & None \\
URL & TabNet & 0.947 & 0.700 & 0.626 & 0.994 & 0.495 & Adversarial & None \\
URL & TabNet & 0.951 & 0.699 & 0.625 & 0.994 & 0.493 & Standard & ctgan \\
URL & TabNet & 0.943 & 0.853 & 0.819 & 0.905 & 0.709 & Adversarial & ctgan \\
URL & TabNet & 0.947 & 0.860 & 0.802 & 0.958 & 0.735 & Standard & cutmix \\
URL & TabNet & 0.935 & 0.860 & 0.815 & 0.934 & 0.729 & Adversarial & cutmix \\
URL & TabNet & 0.934 & 0.851 & 0.803 & 0.932 & 0.712 & Standard & goggle \\
URL & TabNet & 0.939 & 0.868 & 0.880 & 0.852 & 0.736 & Adversarial & goggle \\
URL & TabNet & 0.946 & 0.612 & 0.564 & 0.997 & 0.352 & Standard & wgan \\
URL & TabNet & 0.956 & 0.853 & 0.821 & 0.901 & 0.709 & Adversarial & wgan \\
URL & TabNet & 0.938 & 0.858 & 0.830 & 0.899 & 0.718 & Standard & tablegan \\
URL & TabNet & 0.929 & 0.504 & 1.000 & 0.008 & 0.063 & Adversarial & tablegan \\
URL & TabNet & 0.949 & 0.861 & 0.813 & 0.939 & 0.731 & Standard & tvae \\
URL & TabNet & 0.942 & 0.864 & 0.817 & 0.940 & 0.737 & Adversarial & tvae \\
URL & RLN & 0.984 & 0.945 & 0.945 & 0.946 & 0.891 & Standard & None \\
URL & RLN & 0.977 & 0.933 & 0.917 & 0.953 & 0.867 & Adversarial & None \\
URL & RLN & 0.980 & 0.939 & 0.938 & 0.941 & 0.878 & Standard & ctgan \\
URL & RLN & 0.973 & 0.925 & 0.914 & 0.939 & 0.851 & Adversarial & ctgan \\
URL & RLN & 0.983 & 0.944 & 0.945 & 0.942 & 0.887 & Standard & cutmix \\
URL & RLN & 0.977 & 0.933 & 0.924 & 0.944 & 0.866 & Adversarial & cutmix \\
URL & RLN & 0.978 & 0.938 & 0.937 & 0.940 & 0.877 & Standard & goggle \\
URL & RLN & 0.969 & 0.927 & 0.916 & 0.939 & 0.853 & Adversarial & goggle \\
URL & RLN & 0.982 & 0.940 & 0.945 & 0.934 & 0.880 & Standard & wgan \\
URL & RLN & 0.976 & 0.927 & 0.923 & 0.933 & 0.855 & Adversarial & wgan \\
URL & RLN & 0.980 & 0.934 & 0.953 & 0.913 & 0.868 & Standard & tablegan \\
URL & RLN & 0.971 & 0.925 & 0.933 & 0.915 & 0.850 & Adversarial & tablegan \\
URL & RLN & 0.982 & 0.941 & 0.939 & 0.944 & 0.883 & Standard & tvae \\
URL & RLN & 0.976 & 0.927 & 0.916 & 0.941 & 0.855 & Adversarial & tvae \\
URL & VIME & 0.974 & 0.928 & 0.929 & 0.927 & 0.856 & Standard & None \\
URL & VIME & 0.973 & 0.925 & 0.917 & 0.934 & 0.850 & Adversarial & None \\
URL & VIME & 0.968 & 0.916 & 0.906 & 0.927 & 0.831 & Standard & ctgan \\
URL & VIME & 0.965 & 0.912 & 0.913 & 0.911 & 0.824 & Adversarial & ctgan \\
URL & VIME & 0.971 & 0.922 & 0.921 & 0.924 & 0.844 & Standard & cutmix \\
URL & VIME & 0.967 & 0.918 & 0.915 & 0.921 & 0.836 & Adversarial & cutmix \\
URL & VIME & 0.960 & 0.900 & 0.908 & 0.891 & 0.801 & Standard & goggle \\
URL & VIME & 0.955 & 0.904 & 0.892 & 0.920 & 0.809 & Adversarial & goggle \\
URL & VIME & 0.968 & 0.917 & 0.913 & 0.923 & 0.835 & Standard & wgan \\
URL & VIME & 0.966 & 0.910 & 0.919 & 0.899 & 0.820 & Adversarial & wgan \\
URL & VIME & 0.963 & 0.905 & 0.930 & 0.875 & 0.811 & Standard & tablegan \\
URL & VIME & 0.960 & 0.906 & 0.923 & 0.887 & 0.813 & Adversarial & tablegan \\
URL & VIME & 0.968 & 0.914 & 0.919 & 0.907 & 0.828 & Standard & tvae \\
URL & VIME & 0.964 & 0.908 & 0.915 & 0.899 & 0.816 & Adversarial & tvae \\
LCLD & TabTr & 0.717 & 0.633 & 0.314 & 0.699 & 0.254 & Standard & None \\
LCLD & TabTr & 0.711 & 0.590 & 0.293 & 0.738 & 0.233 & Adversarial & None \\
LCLD & TabTr & 0.711 & 0.614 & 0.304 & 0.715 & 0.244 & Standard & ctgan \\
LCLD & TabTr & 0.694 & 0.526 & 0.271 & 0.803 & 0.212 & Adversarial & ctgan \\
LCLD & TabTr & 0.712 & 0.638 & 0.314 & 0.677 & 0.247 & Standard & cutmix \\
LCLD & TabTr & 0.702 & 0.596 & 0.294 & 0.723 & 0.230 & Adversarial & cutmix \\
LCLD & TabTr & 0.712 & 0.638 & 0.315 & 0.681 & 0.249 & Standard & goggle \\
LCLD & TabTr & 0.699 & 0.645 & 0.312 & 0.636 & 0.231 & Adversarial & goggle \\
LCLD & TabTr & 0.711 & 0.634 & 0.313 & 0.684 & 0.247 & Standard & wgan \\
LCLD & TabTr & 0.688 & 0.615 & 0.296 & 0.664 & 0.214 & Adversarial & wgan \\
LCLD & TabTr & 0.710 & 0.636 & 0.313 & 0.678 & 0.245 & Standard & tablegan \\
LCLD & TabTr & 0.694 & 0.651 & 0.313 & 0.614 & 0.225 & Adversarial & tablegan \\
LCLD & TabTr & 0.716 & 0.634 & 0.314 & 0.693 & 0.252 & Standard & tvae \\
LCLD & TabTr & 0.702 & 0.620 & 0.304 & 0.691 & 0.235 & Adversarial & tvae \\
LCLD & STG & 0.709 & 0.646 & 0.317 & 0.660 & 0.245 & Standard & None \\
LCLD & STG & 0.679 & 0.788 & 0.432 & 0.172 & 0.170 & Adversarial & None \\
LCLD & STG & 0.705 & 0.503 & 0.266 & 0.841 & 0.215 & Standard & ctgan \\
LCLD & STG & 0.700 & 0.505 & 0.266 & 0.833 & 0.212 & Adversarial & ctgan \\
LCLD & STG & 0.707 & 0.766 & 0.404 & 0.347 & 0.231 & Standard & cutmix \\
LCLD & STG & 0.703 & 0.758 & 0.393 & 0.371 & 0.232 & Adversarial & cutmix \\
LCLD & STG & 0.704 & 0.677 & 0.331 & 0.591 & 0.242 & Standard & goggle \\
LCLD & STG & 0.698 & 0.616 & 0.300 & 0.687 & 0.229 & Adversarial & goggle \\
LCLD & STG & 0.705 & 0.669 & 0.326 & 0.606 & 0.241 & Standard & wgan \\
LCLD & STG & 0.699 & 0.657 & 0.318 & 0.617 & 0.234 & Adversarial & wgan \\
LCLD & STG & 0.702 & 0.710 & 0.349 & 0.509 & 0.238 & Standard & tablegan \\
LCLD & STG & 0.699 & 0.657 & 0.318 & 0.621 & 0.235 & Adversarial & tablegan \\
LCLD & STG & 0.706 & 0.652 & 0.319 & 0.645 & 0.244 & Standard & tvae \\
LCLD & STG & 0.706 & 0.625 & 0.307 & 0.687 & 0.239 & Adversarial & tvae \\
LCLD & TabNet & 0.722 & 0.656 & 0.326 & 0.668 & 0.262 & Standard & None \\
LCLD & TabNet & 0.656 & 0.799 & 0.000 & 0.000 & 0.000 & Adversarial & None \\
LCLD & TabNet & 0.687 & 0.785 & 0.270 & 0.042 & 0.031 & Standard & ctgan \\
LCLD & TabNet & 0.695 & 0.799 & 0.000 & 0.000 & 0.000 & Adversarial & ctgan \\
LCLD & TabNet & 0.700 & 0.799 & 1.000 & 0.000 & 0.003 & Standard & cutmix \\
LCLD & TabNet & 0.638 & 0.799 & 0.000 & 0.000 & 0.000 & Adversarial & cutmix \\
LCLD & TabNet & 0.673 & 0.799 & 0.000 & 0.000 & 0.000 & Standard & goggle \\
LCLD & TabNet & 0.683 & 0.201 & 0.201 & 1.000 & 0.000 & Adversarial & goggle \\
LCLD & TabNet & 0.665 & 0.799 & 0.000 & 0.000 & 0.000 & Standard & wgan \\
LCLD & TabNet & 0.688 & 0.799 & 0.000 & 0.000 & 0.000 & Adversarial & wgan \\
LCLD & TabNet & 0.689 & 0.793 & 0.255 & 0.016 & 0.015 & Standard & tablegan \\
LCLD & TabNet & 0.652 & 0.732 & 0.225 & 0.137 & 0.023 & Adversarial & tablegan \\
LCLD & TabNet & 0.667 & 0.799 & 0.248 & 0.000 & 0.002 & Standard & tvae \\
LCLD & TabNet & 0.696 & 0.799 & 0.000 & 0.000 & 0.000 & Adversarial & tvae \\
LCLD & RLN & 0.719 & 0.641 & 0.318 & 0.685 & 0.255 & Standard & None \\
LCLD & RLN & 0.716 & 0.628 & 0.309 & 0.693 & 0.245 & Adversarial & None \\
LCLD & RLN & 0.709 & 0.620 & 0.306 & 0.703 & 0.242 & Standard & ctgan \\
LCLD & RLN & 0.704 & 0.582 & 0.290 & 0.749 & 0.232 & Adversarial & ctgan \\
LCLD & RLN & 0.715 & 0.633 & 0.313 & 0.693 & 0.250 & Standard & cutmix \\
LCLD & RLN & 0.706 & 0.683 & 0.334 & 0.580 & 0.243 & Adversarial & cutmix \\
LCLD & RLN & 0.717 & 0.648 & 0.321 & 0.672 & 0.255 & Standard & goggle \\
LCLD & RLN & 0.710 & 0.644 & 0.317 & 0.666 & 0.247 & Adversarial & goggle \\
LCLD & RLN & 0.712 & 0.644 & 0.317 & 0.668 & 0.248 & Standard & wgan \\
LCLD & RLN & 0.705 & 0.646 & 0.316 & 0.653 & 0.241 & Adversarial & wgan \\
LCLD & RLN & 0.712 & 0.642 & 0.316 & 0.672 & 0.249 & Standard & tablegan \\
LCLD & RLN & 0.704 & 0.629 & 0.308 & 0.679 & 0.239 & Adversarial & tablegan \\
LCLD & RLN & 0.717 & 0.633 & 0.314 & 0.697 & 0.253 & Standard & tvae \\
LCLD & RLN & 0.708 & 0.635 & 0.312 & 0.676 & 0.244 & Adversarial & tvae \\
LCLD & VIME & 0.714 & 0.645 & 0.318 & 0.671 & 0.251 & Standard & None \\
LCLD & VIME & 0.713 & 0.651 & 0.321 & 0.657 & 0.250 & Adversarial & None \\
LCLD & VIME & 0.706 & 0.571 & 0.287 & 0.766 & 0.231 & Standard & ctgan \\
LCLD & VIME & 0.701 & 0.535 & 0.275 & 0.803 & 0.220 & Adversarial & ctgan \\
LCLD & VIME & 0.710 & 0.710 & 0.353 & 0.528 & 0.249 & Standard & cutmix \\
LCLD & VIME & 0.701 & 0.682 & 0.332 & 0.575 & 0.239 & Adversarial & cutmix \\
LCLD & VIME & 0.714 & 0.666 & 0.328 & 0.633 & 0.253 & Standard & goggle \\
LCLD & VIME & 0.703 & 0.685 & 0.334 & 0.569 & 0.239 & Adversarial & goggle \\
LCLD & VIME & 0.708 & 0.648 & 0.318 & 0.658 & 0.247 & Standard & wgan \\
LCLD & VIME & 0.699 & 0.660 & 0.320 & 0.618 & 0.237 & Adversarial & wgan \\
LCLD & VIME & 0.708 & 0.676 & 0.332 & 0.606 & 0.249 & Standard & tablegan \\
LCLD & VIME & 0.696 & 0.677 & 0.327 & 0.574 & 0.232 & Adversarial & tablegan \\
LCLD & VIME & 0.714 & 0.654 & 0.322 & 0.657 & 0.252 & Standard & tvae \\
LCLD & VIME & 0.705 & 0.628 & 0.308 & 0.684 & 0.240 & Adversarial & tvae \\
CTU & TabTr & 0.979 & 1.000 & 0.982 & 0.953 & 0.967 & Standard & None \\
CTU & TabTr & 0.985 & 1.000 & 0.982 & 0.953 & 0.967 & Adversarial & None \\
CTU & TabTr & 0.630 & 0.044 & 0.008 & 1.000 & 0.017 & Standard & ctgan \\
CTU & TabTr & 0.627 & 0.045 & 0.008 & 1.000 & 0.017 & Adversarial & ctgan \\
CTU & TabTr & 0.977 & 1.000 & 0.982 & 0.953 & 0.967 & Standard & cutmix \\
CTU & TabTr & 0.980 & 1.000 & 0.982 & 0.953 & 0.967 & Adversarial & cutmix \\
CTU & TabTr & 0.982 & 1.000 & 0.982 & 0.953 & 0.967 & Standard & wgan \\
CTU & TabTr & 0.984 & 1.000 & 0.982 & 0.953 & 0.967 & Adversarial & wgan \\
CTU & TabTr & 0.978 & 1.000 & 0.987 & 0.951 & 0.969 & Standard & tablegan \\
CTU & TabTr & 0.979 & 1.000 & 0.987 & 0.953 & 0.970 & Adversarial & tablegan \\
CTU & TabTr & 0.977 & 0.943 & 0.111 & 0.963 & 0.317 & Standard & tvae \\
CTU & TabTr & 0.974 & 0.609 & 0.018 & 0.983 & 0.103 & Adversarial & tvae \\
CTU & STG & 0.988 & 1.000 & 0.982 & 0.953 & 0.967 & Standard & None \\
CTU & STG & 0.986 & 1.000 & 0.992 & 0.951 & 0.971 & Adversarial & None \\
CTU & STG & 0.990 & 0.999 & 0.890 & 0.956 & 0.922 & Standard & ctgan \\
CTU & STG & 0.986 & 0.930 & 0.092 & 0.961 & 0.286 & Adversarial & ctgan \\
CTU & STG & 0.986 & 1.000 & 0.982 & 0.953 & 0.967 & Standard & cutmix \\
CTU & STG & 0.985 & 1.000 & 1.000 & 0.946 & 0.972 & Adversarial & cutmix \\
CTU & STG & 0.986 & 1.000 & 0.982 & 0.953 & 0.967 & Standard & wgan \\
CTU & STG & 0.985 & 1.000 & 0.982 & 0.953 & 0.967 & Adversarial & wgan \\
CTU & STG & 0.986 & 1.000 & 0.982 & 0.953 & 0.967 & Standard & tablegan \\
CTU & STG & 0.984 & 1.000 & 1.000 & 0.951 & 0.975 & Adversarial & tablegan \\
CTU & STG & 0.984 & 0.890 & 0.061 & 0.963 & 0.227 & Standard & tvae \\
CTU & STG & 0.981 & 0.436 & 0.013 & 0.983 & 0.072 & Adversarial & tvae \\
CTU & TabNet & 0.996 & 0.999 & 0.958 & 0.961 & 0.959 & Standard & None \\
CTU & TabNet & 0.978 & 0.993 & 0.500 & 0.002 & 0.035 & Adversarial & None \\
CTU & TabNet & 0.986 & 0.993 & 0.000 & 0.000 & 0.000 & Standard & ctgan \\
CTU & TabNet & 0.977 & 0.016 & 0.007 & 1.000 & 0.008 & Adversarial & ctgan \\
CTU & TabNet & 0.982 & 0.993 & 0.000 & 0.000 & 0.000 & Standard & cutmix \\
CTU & TabNet & 0.982 & 0.993 & 0.000 & 0.000 & 0.000 & Adversarial & cutmix \\
CTU & TabNet & 0.983 & 1.000 & 0.985 & 0.951 & 0.967 & Standard & wgan \\
CTU & TabNet & 0.987 & 0.993 & 0.000 & 0.000 & 0.000 & Adversarial & wgan \\
CTU & TabNet & 0.980 & 1.000 & 0.982 & 0.953 & 0.967 & Standard & tablegan \\
CTU & TabNet & 0.993 & 0.993 & 1.000 & 0.015 & 0.121 & Adversarial & tablegan \\
CTU & TabNet & 0.987 & 0.993 & 0.000 & 0.000 & 0.000 & Standard & tvae \\
CTU & TabNet & 0.976 & 0.007 & 0.007 & 1.000 & 0.000 & Adversarial & tvae \\
CTU & RLN & 0.991 & 0.998 & 0.819 & 0.978 & 0.894 & Standard & None \\
CTU & RLN & 0.990 & 0.999 & 0.904 & 0.973 & 0.937 & Adversarial & None \\
CTU & RLN & 0.994 & 0.986 & 0.338 & 0.975 & 0.570 & Standard & ctgan \\
CTU & RLN & 0.992 & 0.985 & 0.327 & 0.975 & 0.561 & Adversarial & ctgan \\
CTU & RLN & 0.989 & 1.000 & 0.987 & 0.953 & 0.970 & Standard & cutmix \\
CTU & RLN & 0.987 & 1.000 & 1.000 & 0.953 & 0.976 & Adversarial & cutmix \\
CTU & RLN & 0.991 & 0.999 & 0.887 & 0.966 & 0.925 & Standard & wgan \\
CTU & RLN & 0.990 & 0.999 & 0.923 & 0.975 & 0.949 & Adversarial & wgan \\
CTU & RLN & 0.992 & 0.999 & 0.880 & 0.975 & 0.926 & Standard & tablegan \\
CTU & RLN & 0.990 & 0.999 & 0.896 & 0.975 & 0.934 & Adversarial & tablegan \\
CTU & RLN & 0.988 & 0.987 & 0.362 & 0.973 & 0.589 & Standard & tvae \\
CTU & RLN & 0.988 & 0.986 & 0.338 & 0.975 & 0.570 & Adversarial & tvae \\
CTU & VIME & 0.987 & 1.000 & 0.997 & 0.951 & 0.974 & Standard & None \\
CTU & VIME & 0.983 & 1.000 & 0.997 & 0.951 & 0.974 & Adversarial & None \\
CTU & VIME & 0.972 & 0.007 & 0.007 & 1.000 & 0.000 & Standard & ctgan \\
CTU & VIME & 0.741 & 0.007 & 0.007 & 1.000 & 0.000 & Adversarial & ctgan \\
CTU & VIME & 0.991 & 1.000 & 0.997 & 0.951 & 0.974 & Standard & cutmix \\
CTU & VIME & 0.976 & 1.000 & 0.997 & 0.951 & 0.974 & Adversarial & cutmix \\
CTU & VIME & 0.977 & 1.000 & 1.000 & 0.953 & 0.976 & Standard & wgan \\
CTU & VIME & 0.979 & 1.000 & 0.997 & 0.953 & 0.975 & Adversarial & wgan \\
CTU & VIME & 0.984 & 1.000 & 0.997 & 0.951 & 0.974 & Standard & tablegan \\
CTU & VIME & 0.979 & 1.000 & 0.997 & 0.951 & 0.974 & Adversarial & tablegan \\
CTU & VIME & 0.950 & 0.008 & 0.007 & 1.000 & 0.001 & Standard & tvae \\
CTU & VIME & 0.727 & 0.007 & 0.007 & 1.000 & 0.000 & Adversarial & tvae \\
WIDS & TabTr & 0.874 & 0.810 & 0.287 & 0.755 & 0.383 & Standard & None \\
WIDS & TabTr & 0.869 & 0.794 & 0.272 & 0.772 & 0.373 & Adversarial & None \\
WIDS & TabTr & 0.868 & 0.799 & 0.279 & 0.780 & 0.383 & Standard & ctgan \\
WIDS & TabTr & 0.859 & 0.769 & 0.249 & 0.782 & 0.349 & Adversarial & ctgan \\
WIDS & TabTr & 0.866 & 0.835 & 0.314 & 0.708 & 0.395 & Standard & cutmix \\
WIDS & TabTr & 0.851 & 0.867 & 0.358 & 0.601 & 0.395 & Adversarial & cutmix \\
WIDS & TabTr & 0.873 & 0.805 & 0.285 & 0.784 & 0.392 & Standard & goggle \\
WIDS & TabTr & 0.853 & 0.784 & 0.261 & 0.764 & 0.357 & Adversarial & goggle \\
WIDS & TabTr & 0.866 & 0.797 & 0.273 & 0.763 & 0.371 & Standard & wgan \\
WIDS & TabTr & 0.864 & 0.788 & 0.264 & 0.764 & 0.361 & Adversarial & wgan \\
WIDS & TabTr & 0.869 & 0.808 & 0.284 & 0.748 & 0.378 & Standard & tablegan \\
WIDS & TabTr & 0.858 & 0.806 & 0.277 & 0.724 & 0.363 & Adversarial & tablegan \\
WIDS & TabTr & 0.871 & 0.801 & 0.280 & 0.776 & 0.383 & Standard & tvae \\
WIDS & TabTr & 0.858 & 0.790 & 0.264 & 0.747 & 0.356 & Adversarial & tvae \\
WIDS & STG & 0.866 & 0.782 & 0.260 & 0.776 & 0.361 & Standard & None \\
WIDS & STG & 0.865 & 0.875 & 0.381 & 0.627 & 0.424 & Adversarial & None \\
WIDS & STG & 0.852 & 0.638 & 0.183 & 0.878 & 0.285 & Standard & ctgan \\
WIDS & STG & 0.841 & 0.668 & 0.193 & 0.851 & 0.293 & Adversarial & ctgan \\
WIDS & STG & 0.863 & 0.885 & 0.400 & 0.567 & 0.414 & Standard & cutmix \\
WIDS & STG & 0.851 & 0.880 & 0.380 & 0.530 & 0.385 & Adversarial & cutmix \\
WIDS & STG & 0.851 & 0.780 & 0.253 & 0.742 & 0.342 & Standard & goggle \\
WIDS & STG & 0.837 & 0.727 & 0.218 & 0.787 & 0.310 & Adversarial & goggle \\
WIDS & STG & 0.863 & 0.800 & 0.274 & 0.744 & 0.366 & Standard & wgan \\
WIDS & STG & 0.855 & 0.855 & 0.334 & 0.625 & 0.384 & Adversarial & wgan \\
WIDS & STG & 0.861 & 0.846 & 0.326 & 0.676 & 0.396 & Standard & tablegan \\
WIDS & STG & 0.853 & 0.829 & 0.302 & 0.688 & 0.376 & Adversarial & tablegan \\
WIDS & STG & 0.857 & 0.776 & 0.252 & 0.758 & 0.345 & Standard & tvae \\
WIDS & STG & 0.845 & 0.807 & 0.271 & 0.678 & 0.341 & Adversarial & tvae \\
WIDS & TabNet & 0.870 & 0.777 & 0.259 & 0.796 & 0.365 & Standard & None \\
WIDS & TabNet & 0.835 & 0.104 & 0.090 & 0.984 & 0.003 & Adversarial & None \\
WIDS & TabNet & 0.853 & 0.090 & 0.090 & 1.000 & 0.000 & Standard & ctgan \\
WIDS & TabNet & 0.863 & 0.090 & 0.090 & 1.000 & 0.000 & Adversarial & ctgan \\
WIDS & TabNet & 0.866 & 0.910 & 0.000 & 0.000 & 0.000 & Standard & cutmix \\
WIDS & TabNet & 0.859 & 0.090 & 0.090 & 1.000 & 0.000 & Adversarial & cutmix \\
WIDS & TabNet & 0.856 & 0.090 & 0.090 & 1.000 & 0.000 & Standard & goggle \\
WIDS & TabNet & 0.862 & 0.090 & 0.090 & 1.000 & 0.000 & Adversarial & goggle \\
WIDS & TabNet & 0.865 & 0.795 & 0.275 & 0.787 & 0.381 & Standard & wgan \\
WIDS & TabNet & 0.855 & 0.090 & 0.090 & 1.000 & 0.000 & Adversarial & wgan \\
WIDS & TabNet & 0.864 & 0.090 & 0.090 & 1.000 & 0.000 & Standard & tablegan \\
WIDS & TabNet & 0.860 & 0.090 & 0.090 & 1.000 & 0.000 & Adversarial & tablegan \\
WIDS & TabNet & 0.857 & 0.104 & 0.090 & 0.984 & 0.003 & Standard & tvae \\
WIDS & TabNet & 0.864 & 0.090 & 0.090 & 1.000 & 0.000 & Adversarial & tvae \\
WIDS & RLN & 0.869 & 0.796 & 0.274 & 0.774 & 0.376 & Standard & None \\
WIDS & RLN & 0.867 & 0.789 & 0.268 & 0.779 & 0.370 & Adversarial & None \\
WIDS & RLN & 0.862 & 0.788 & 0.264 & 0.761 & 0.360 & Standard & ctgan \\
WIDS & RLN & 0.425 & 0.090 & 0.090 & 1.000 & 0.000 & Adversarial & ctgan \\
WIDS & RLN & 0.870 & 0.802 & 0.280 & 0.769 & 0.381 & Standard & cutmix \\
WIDS & RLN & 0.859 & 0.834 & 0.307 & 0.681 & 0.379 & Adversarial & cutmix \\
WIDS & RLN & 0.864 & 0.797 & 0.276 & 0.774 & 0.378 & Standard & goggle \\
WIDS & RLN & 0.857 & 0.777 & 0.256 & 0.782 & 0.358 & Adversarial & goggle \\
WIDS & RLN & 0.866 & 0.782 & 0.260 & 0.774 & 0.359 & Standard & wgan \\
WIDS & RLN & 0.858 & 0.770 & 0.249 & 0.776 & 0.347 & Adversarial & wgan \\
WIDS & RLN & 0.868 & 0.773 & 0.254 & 0.785 & 0.356 & Standard & tablegan \\
WIDS & RLN & 0.860 & 0.797 & 0.273 & 0.760 & 0.370 & Adversarial & tablegan \\
WIDS & RLN & 0.868 & 0.776 & 0.259 & 0.803 & 0.367 & Standard & tvae \\
WIDS & RLN & 0.854 & 0.756 & 0.237 & 0.774 & 0.332 & Adversarial & tvae \\
WIDS & VIME & 0.865 & 0.823 & 0.298 & 0.721 & 0.384 & Standard & None \\
WIDS & VIME & 0.858 & 0.817 & 0.291 & 0.720 & 0.376 & Adversarial & None \\
WIDS & VIME & 0.482 & 0.090 & 0.090 & 1.000 & 0.000 & Standard & ctgan \\
WIDS & VIME & 0.482 & 0.090 & 0.090 & 1.000 & 0.000 & Adversarial & ctgan \\
WIDS & VIME & 0.857 & 0.833 & 0.309 & 0.697 & 0.387 & Standard & cutmix \\
WIDS & VIME & 0.849 & 0.878 & 0.374 & 0.543 & 0.385 & Adversarial & cutmix \\
WIDS & VIME & 0.849 & 0.812 & 0.280 & 0.700 & 0.358 & Standard & goggle \\
WIDS & VIME & 0.840 & 0.802 & 0.268 & 0.700 & 0.346 & Adversarial & goggle \\
WIDS & VIME & 0.861 & 0.796 & 0.270 & 0.753 & 0.365 & Standard & wgan \\
WIDS & VIME & 0.845 & 0.791 & 0.259 & 0.715 & 0.339 & Adversarial & wgan \\
WIDS & VIME & 0.864 & 0.828 & 0.305 & 0.716 & 0.389 & Standard & tablegan \\
WIDS & VIME & 0.853 & 0.882 & 0.388 & 0.553 & 0.399 & Adversarial & tablegan \\
WIDS & VIME & 0.858 & 0.808 & 0.280 & 0.726 & 0.367 & Standard & tvae \\
WIDS & VIME & 0.846 & 0.787 & 0.256 & 0.721 & 0.339 & Adversarial & tvae \\
\bottomrule

\end{longtable}

For LCLD dataset only Goggle and WGAN data augmentations lead to $MCC=0$. To uncover what happens with some generated data, we study the distribution of artificial examples on the LCLD dataset for 3 cases: Two cases where performance did not collapse: TableGAN and CTGAN and one problematic case WGAN.

\begin{figure}[!h]
\centering
\begin{subfigure}{0.45\textwidth}
    \includegraphics[clip, width=\textwidth]{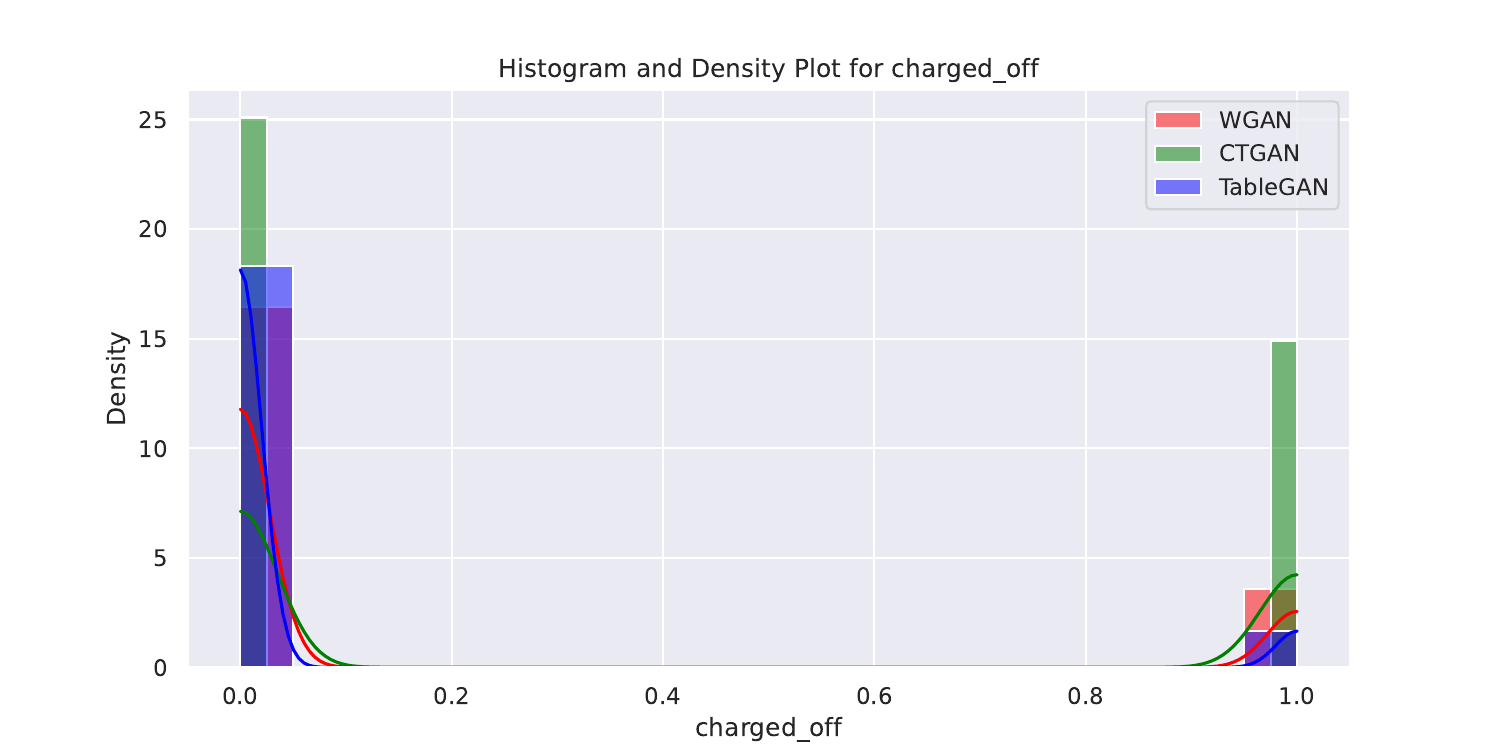}
    \caption{Label: Charged off}
    \label{fig:app-kde-charged}
\end{subfigure}
\begin{subfigure}{0.45\textwidth}
    \includegraphics[clip, width=\textwidth]{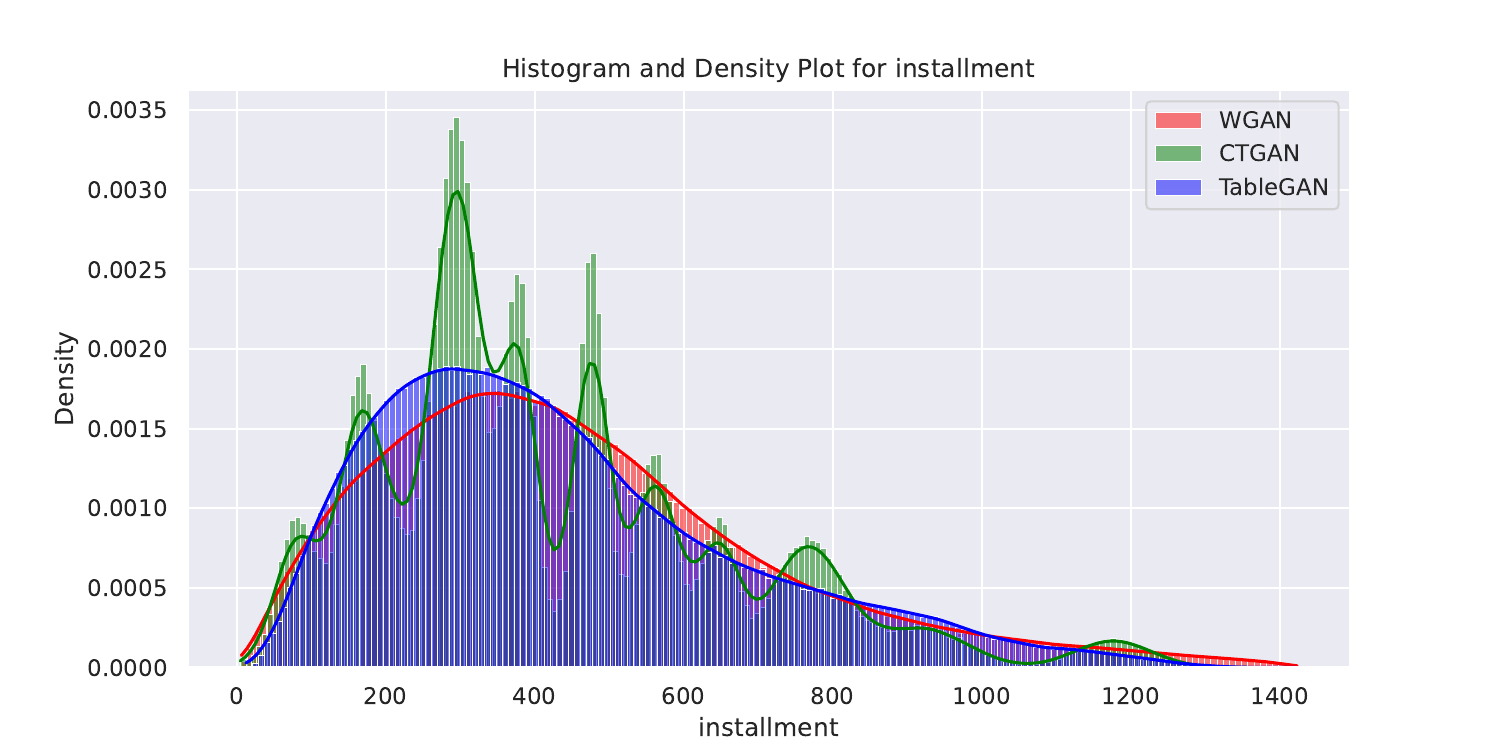}
    \caption{Feature: Installment}
    \label{fig:app-kde-install}
\end{subfigure}
\hfill
\begin{subfigure}{0.45\textwidth}
    \includegraphics[clip, width=\textwidth]{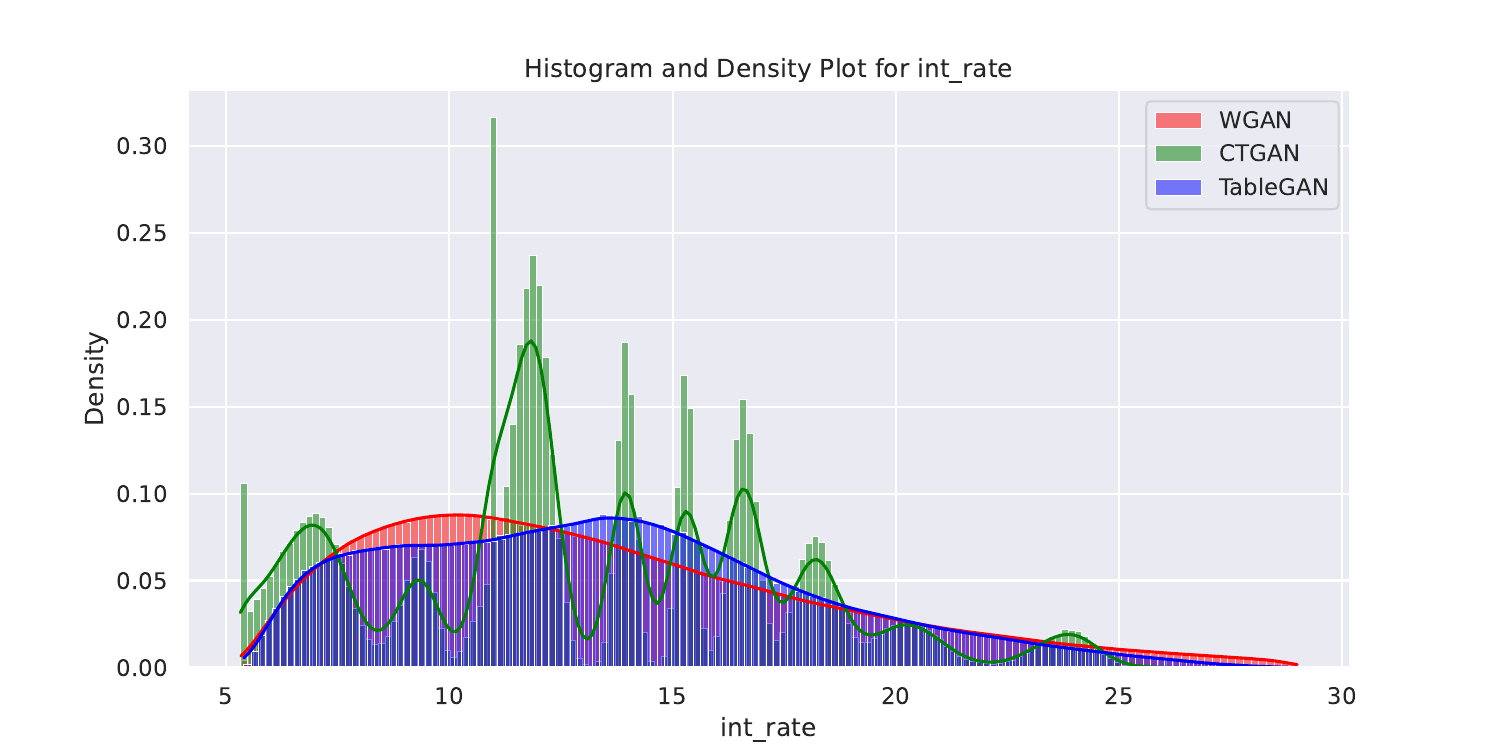}
    \caption{Feature: Interest rate}
    \label{fig:app-kde-rate}
\end{subfigure}
\begin{subfigure}{0.45\textwidth}
    \includegraphics[clip, width=\textwidth]{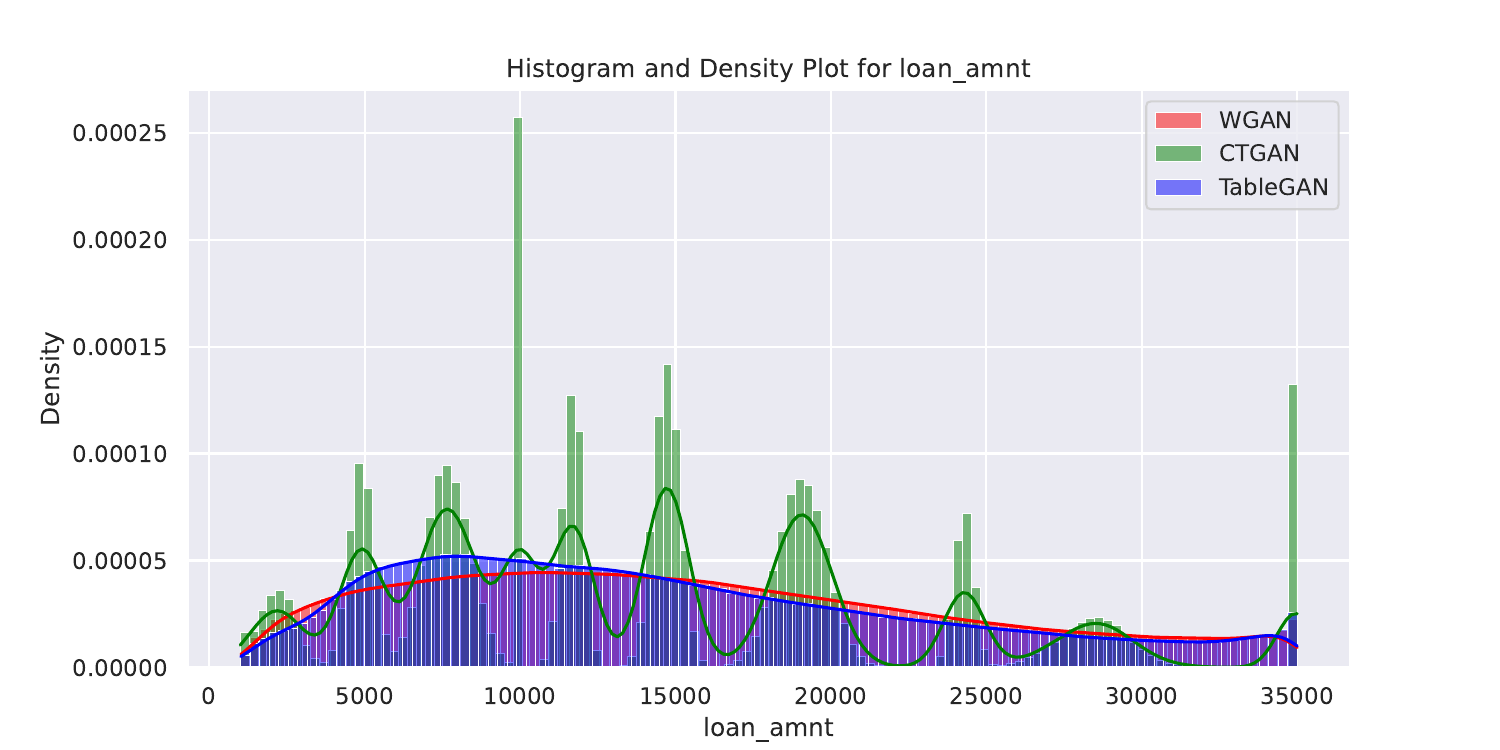}
    \caption{Feature: Loan amount}
    \label{fig:app-kde-amnt}
\end{subfigure}
\hfill
\begin{subfigure}{0.45\textwidth}
    \includegraphics[clip,width=\textwidth]{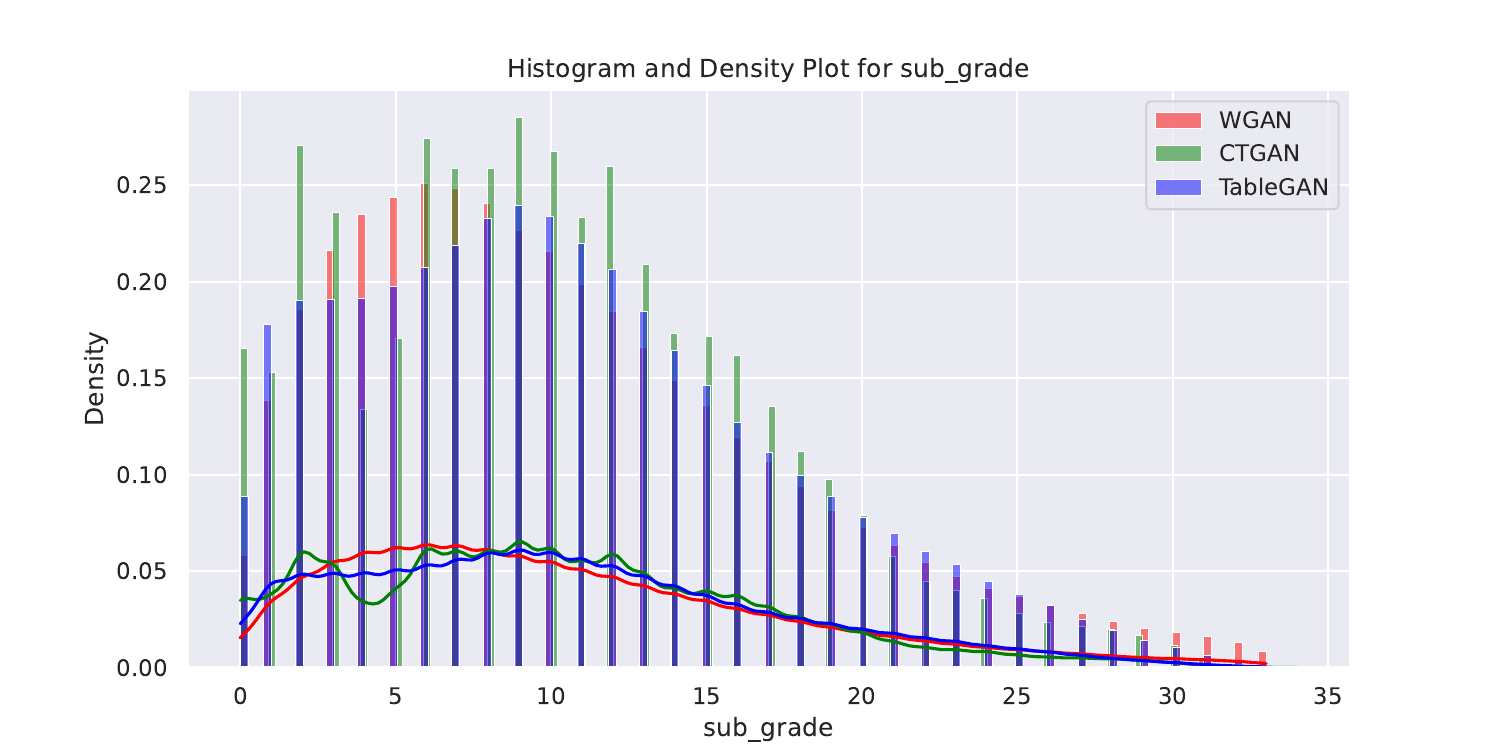}
    \caption{Feature: Sub-grade}
    \label{fig:app-kde-grade}
\end{subfigure}
\begin{subfigure}{0.45\textwidth}
    \includegraphics[clip,width=\textwidth]{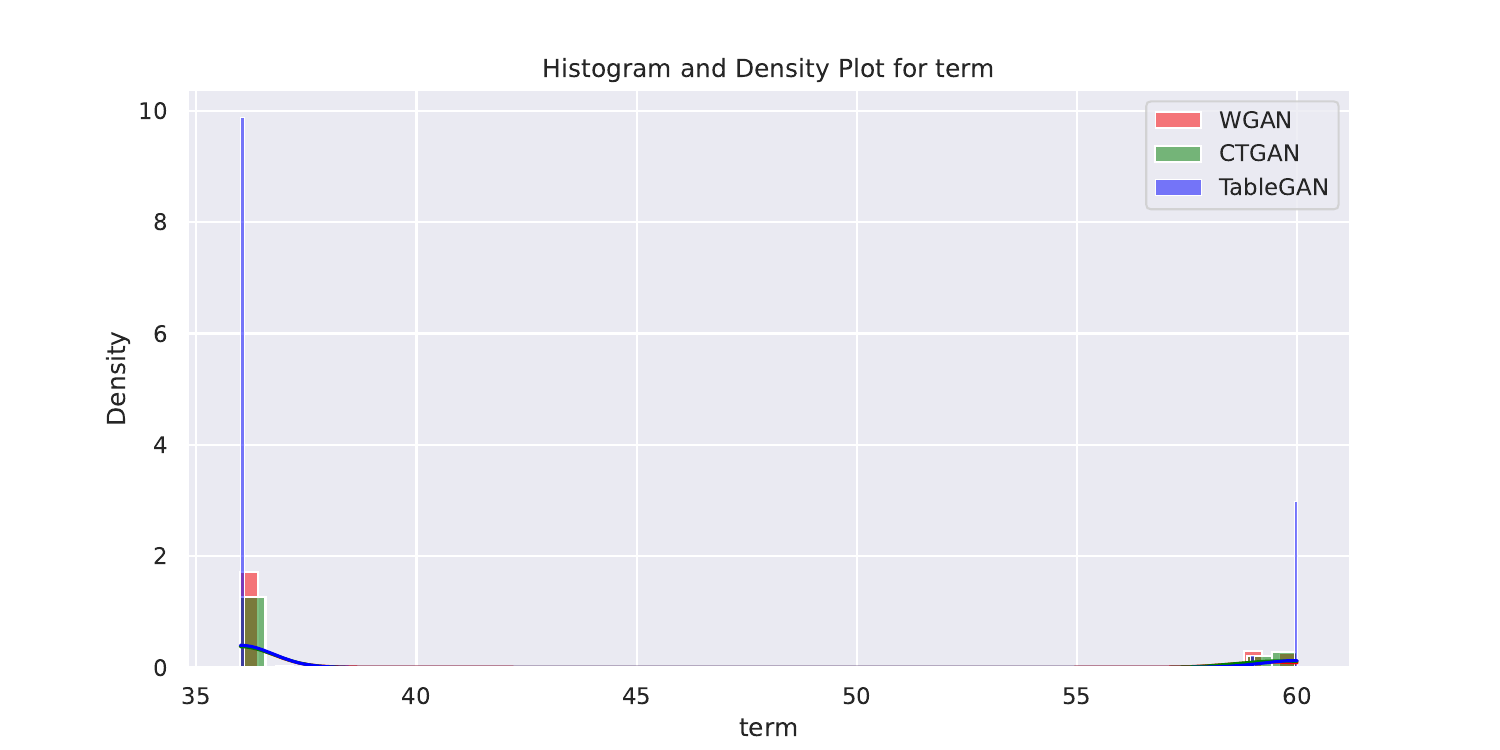}
    \caption{Feature: Loan term}
    \label{fig:app-kde-term}
\end{subfigure}

\caption{Impact of attack budget on the robust accuracy for LCLD dataset.}
\label{fig:appendix-kde}
\vspace{-1.5em}
\end{figure}

\paragraph{Kernel Density Estimation.} We first compare the artificial examples distributions in Figure \ref{fig:appendix-kde}. The results show that the labels and the main features of TableGAN, a "healthy" generator are closer to the distribution of the "problematic" generator WGAN than to the distribution of CTGAN, another "healthy" generator. Feature and label distributions are not problematic.

\paragraph{Statistical analysis.} 

We perform the following statistical tests to compare the distributions quantitatively between the examples generated by the three generators. Kolmogorov-Smirnov test, t-test, or MWU test. We report the results in Table \ref{tab:app-stat-test}. %
Across all statistical tests, there is no specific pattern to the faulty generator "WGAN" compared to CTGAN and TableGAN.

\begin{table}[!h]
    \centering
     \caption{Statistical tests between the distributions of the 3 generators: W:WGAN, T:TableGAN, C:CTGAN, MWU:Mann-Whitney U }
    \label{tab:app-stat-test}
    
    \begin{tabular}{l lllllll}
\toprule
GAN & Test & Amount & Term & Rate & Installment & Sub-grade & Label \\
\midrule
(W/T)  & KS Statistic & 0.047 & 0.120 & 0.055 & 0.046 & 0.031 & 0.095 \\
(W/T)  & KS p-value & 0.000 & 0.000 & 0.000 & 0.000 & 0.000 & 0.000 \\
\midrule
(W/T)  & t-test Statistic & 35.923 & 10.782 & -7.687 & 40.512 & 0.224 & 140.654 \\
(W/T) & t-test p-value & 0.000 & 0.000 & 0.000 & 0.000 & 0.823 & 0.000 \\
\midrule
(W/T)  & MWU Statistic & \num[round-precision=2,round-mode=figures,
     scientific-notation=true]{127570297453.500} & \num[round-precision=2,round-mode=figures,
     scientific-notation=true]{124021831892.000} & \num[round-precision=2,round-mode=figures,
     scientific-notation=true]{118060888498.000} & \num[round-precision=2,round-mode=figures,
     scientific-notation=true]{127691200271.000} & \num[round-precision=2,round-mode=figures,
     scientific-notation=true]{120535248445.000} & \num[round-precision=2,round-mode=figures,
     scientific-notation=true]{133607818344.000} \\
(W/T)  & MWU p-value & 0.000 & 0.000 & 0.000 & 0.000 & 0.000 & 0.000 \\
\midrule
\midrule
(W/C)  &  KS Statistic & 0.112 & 0.056 & 0.105 & 0.089 & 0.037 & 0.194 \\
(W/C)  &  KS p-value & 0.000 & 0.000 & 0.000 & 0.000 & 0.000 & 0.000 \\
\midrule
(W/C)  &  t-test Statistic & 80.112 & -21.286 & 40.896 & 61.097 & 30.043 & -221.351 \\
(W/C)  & t-test p-value & 0.000 & 0.000 & 0.000 & 0.000 & 0.000 & 0.000 \\
\midrule
(W/C)  &  MWU Statistic & \num[round-precision=2,round-mode=figures,
     scientific-notation=true]{133208951910.500} & \num[round-precision=2,round-mode=figures,
     scientific-notation=true]{122417395219.000} & \num[round-precision=2,round-mode=figures,
     scientific-notation=true]{123425726352.000} & \num[round-precision=2,round-mode=figures,
     scientific-notation=true]{130353018289.000} & \num[round-precision=2,round-mode=figures,
     scientific-notation=true]{123732783848.000} & \num[round-precision=2,round-mode=figures,
     scientific-notation=true]{98353651368.000} \\
(W/C)  &  MWU p-value & 0.000 & 0.002 & 0.000 & 0.000 & 0.000 & 0.000 \\
\midrule
\midrule
(T/C)  & KS Statistic & 0.079 & 0.070 & 0.093 & 0.044 & 0.027 & 0.289 \\
(T/C)  & KS p-value & 0.000 & 0.000 & 0.000 & 0.000 & 0.000 & 0.000 \\
\midrule
(T/C)  & t-test Statistic & -43.986 & 31.467 & -51.028 & -20.991 & -30.376 & 364.250 \\
(T/C)  & t-test p-value & 0.000 & 0.000 & 0.000 & 0.000 & 0.000 & 0.000 \\
\midrule
(T/C)  & MWU Statistic & \num[round-precision=2,round-mode=figures,
     scientific-notation=true]{116617560129.000} & \num[round-precision=2,round-mode=figures,
     scientific-notation=true]{122646460321.000} & \num[round-precision=2,round-mode=figures,
     scientific-notation=true]{116284769942.500} & \num[round-precision=2,round-mode=figures,
     scientific-notation=true]{119431245351.500} & \num[round-precision=2,round-mode=figures,
     scientific-notation=true]{118729891920.000} & \num[round-precision=2,round-mode=figures,
     scientific-notation=true]{157315642848.000} \\
(T/C)  & MWU p-value & 0.000 & 0.000 & 0.000 & 0.000 & 0.000 & 0.000 \\
\bottomrule
\end{tabular}

\end{table}

\paragraph{Classification performance.}

We build a new classifier to identify examples generated by WGAN and by TableGAN. 
We leverage Oodeel\footnote{https://github.com/deel-ai/oodeel}, a library that performs post-hoc deep OOD (Out-of-Distribution) detection.

The classifier reaches achieves a random accuracy (0.5) confirming that no specific features are sufficient to distinguish both generators.

\begin{figure}[!h]
\centering
\begin{subfigure}{0.45\textwidth}
    \includegraphics[clip, width=\textwidth]{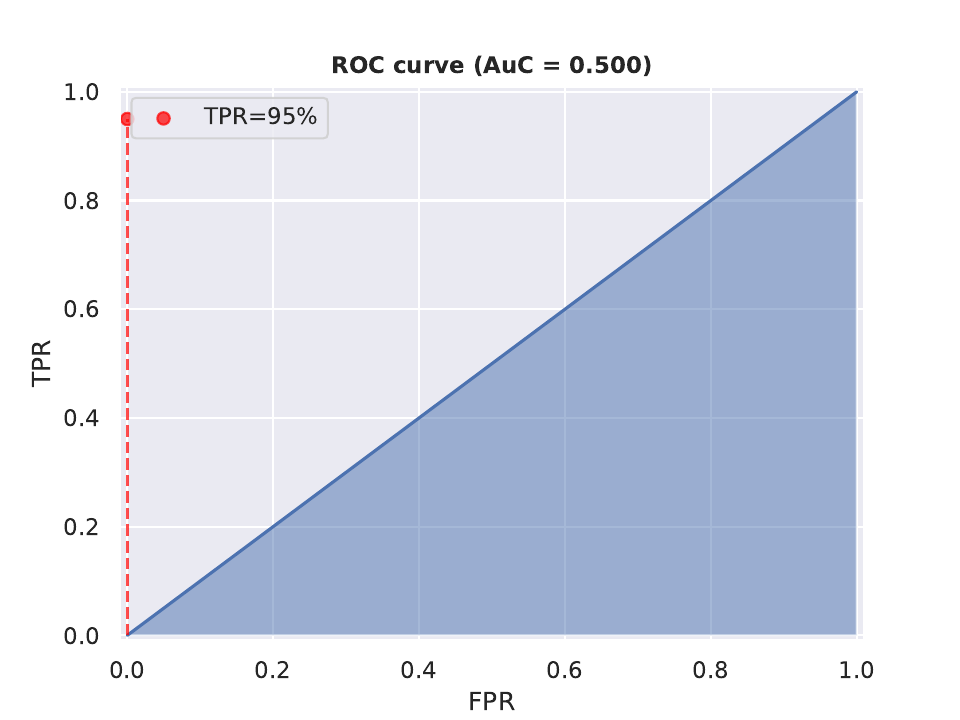}
    \caption{AUC-ROC curve for the OOD detector}
    \label{fig:app-auc-ood}
\end{subfigure}
\begin{subfigure}{0.45\textwidth}
    \includegraphics[clip, width=\textwidth]{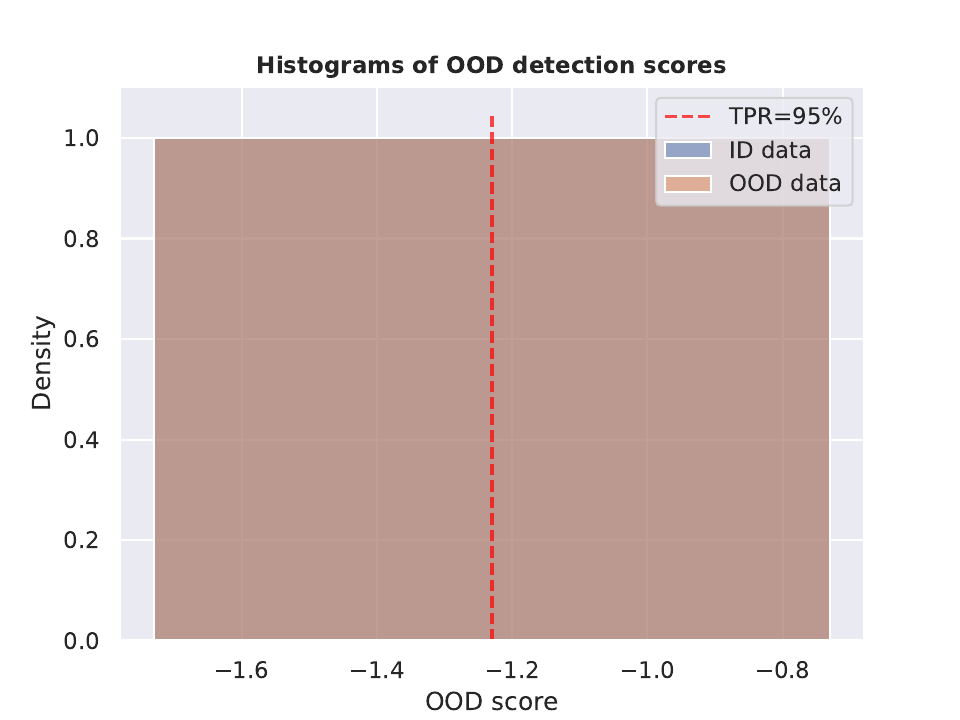}
    \caption{Histograms of OOD detection scores}
    \label{fig:app-hist-ood}
\end{subfigure}

\caption{Performance of the OOD detector on the WGAN samples.}
\label{fig:appendix-detector}
\vspace{-1.5em}
\end{figure}

Next, we evaluate the Maximum Logit Score (MLS) detector and report the histograms and AUROC curve of the detector in Figure \ref{fig:appendix-detector}.

Both the ROC curves and the histograms confirm that WGAN and TableGAN are not distinguishible.

\paragraph{Conclusion:} From all our analysis, we confirm that the collapse of performance of training with WGAN data augmentation is not due to some evident properties in the generated examples.

\paragraph{Robust performance after data augmentation}
We report below the robustness of our 270 models trained with various combinations of arhcitecture, data augmentation, and adversarial training.

\small{
\begin{longtable}{llllrrrrrr}

\caption{Detailed results of adversarial robustness with constrained (CTR) and unconstrained attacks (ADV) across our 5 seeds  }
\label{tab:detailed_robustness_agumentation} \\
Dataset & Arch & Training & Augment & ID$_{mean}$ & CTR$_{mean}$ & ADV$_{mean}$ & ID$_{std}$ & CTR$_{std}$ & ADV$_{std}$ \\
\midrule
CTU & STG & Adversarial & None & 0.951 & 0.951 & 0.951 & 0.000 & 0.000 & 0.000 \\
CTU & STG & Adversarial & ctgan & 0.961 & 0.960 & 0.959 & 0.000 & 0.001 & 0.002 \\
CTU & STG & Adversarial & cutmix & 0.946 & 0.945 & 0.946 & 0.000 & 0.001 & 0.000 \\
CTU & STG & Adversarial & tablegan & 0.951 & 0.951 & 0.951 & 0.000 & 0.000 & 0.000 \\
CTU & STG & Adversarial & tvae & 0.983 & 0.983 & 0.982 & 0.000 & 0.000 & 0.001 \\
CTU & STG & Adversarial & wgan & 0.953 & 0.953 & 0.953 & 0.000 & 0.000 & 0.000 \\
CTU & STG & Standard & None & 0.953 & 0.953 & 0.953 & 0.000 & 0.000 & 0.000 \\
CTU & STG & Standard & ctgan & 0.956 & 0.953 & 0.956 & 0.000 & 0.000 & 0.000 \\
CTU & STG & Standard & cutmix & 0.953 & 0.953 & 0.953 & 0.000 & 0.000 & 0.000 \\
CTU & STG & Standard & tablegan & 0.953 & 0.953 & 0.953 & 0.000 & 0.000 & 0.000 \\
CTU & STG & Standard & tvae & 0.963 & 0.961 & 0.963 & 0.000 & 0.000 & 0.000 \\
CTU & STG & Standard & wgan & 0.953 & 0.953 & 0.953 & 0.000 & 0.000 & 0.000 \\
CTU & TabNet & Adversarial & None & 0.002 & 0.002 & 0.002 & 0.000 & 0.001 & 0.001 \\
CTU & TabNet & Adversarial & ctgan & 1.000 & 1.000 & 1.000 & 0.000 & 0.000 & 0.000 \\
CTU & TabNet & Adversarial & cutmix & 0.000 & 0.000 & 0.000 & 0.000 & 0.000 & 0.000 \\
CTU & TabNet & Adversarial & tablegan & 0.015 & 0.014 & 0.014 & 0.000 & 0.001 & 0.001 \\
CTU & TabNet & Adversarial & tvae & 1.000 & 1.000 & 1.000 & 0.000 & 0.000 & 0.000 \\
CTU & TabNet & Adversarial & wgan & 0.000 & 0.000 & 0.000 & 0.000 & 0.000 & 0.000 \\
CTU & TabNet & Standard & None & 0.961 & 0.000 & 0.961 & 0.000 & 0.000 & 0.000 \\
CTU & TabNet & Standard & ctgan & 0.000 & 0.000 & 0.000 & 0.000 & 0.000 & 0.000 \\
CTU & TabNet & Standard & cutmix & 0.000 & 0.000 & 0.000 & 0.000 & 0.000 & 0.000 \\
CTU & TabNet & Standard & tablegan & 0.953 & 0.953 & 0.953 & 0.000 & 0.000 & 0.000 \\
CTU & TabNet & Standard & tvae & 0.000 & 0.000 & 0.000 & 0.000 & 0.000 & 0.000 \\
CTU & TabNet & Standard & wgan & 0.951 & 0.951 & 0.951 & 0.000 & 0.000 & 0.000 \\
CTU & TabTr & Adversarial & None & 0.953 & 0.953 & 0.953 & 0.000 & 0.000 & 0.000 \\
CTU & TabTr & Adversarial & ctgan & 1.000 & 0.944 & 1.000 & 0.000 & 0.010 & 0.000 \\
CTU & TabTr & Adversarial & cutmix & 0.953 & 0.953 & 0.953 & 0.000 & 0.000 & 0.000 \\
CTU & TabTr & Adversarial & tablegan & 0.953 & 0.953 & 0.953 & 0.000 & 0.001 & 0.000 \\
CTU & TabTr & Adversarial & tvae & 0.983 & 0.983 & 0.983 & 0.000 & 0.000 & 0.000 \\
CTU & TabTr & Adversarial & wgan & 0.953 & 0.953 & 0.953 & 0.000 & 0.000 & 0.000 \\
CTU & TabTr & Standard & None & 0.953 & 0.953 & 0.953 & 0.000 & 0.000 & 0.000 \\
CTU & TabTr & Standard & ctgan & 1.000 & 0.944 & 1.000 & 0.000 & 0.005 & 0.000 \\
CTU & TabTr & Standard & cutmix & 0.953 & 0.949 & 0.953 & 0.000 & 0.003 & 0.000 \\
CTU & TabTr & Standard & tablegan & 0.951 & 0.939 & 0.951 & 0.000 & 0.001 & 0.000 \\
CTU & TabTr & Standard & tvae & 0.963 & 0.961 & 0.963 & 0.000 & 0.000 & 0.000 \\
CTU & TabTr & Standard & wgan & 0.953 & 0.953 & 0.953 & 0.000 & 0.000 & 0.000 \\
CTU & RLN & Adversarial & None & 0.973 & 0.971 & 0.973 & 0.000 & 0.000 & 0.000 \\
CTU & RLN & Adversarial & ctgan & 0.975 & 0.967 & 0.975 & 0.000 & 0.001 & 0.000 \\
CTU & RLN & Adversarial & cutmix & 0.953 & 0.953 & 0.953 & 0.000 & 0.000 & 0.000 \\
CTU & RLN & Adversarial & tablegan & 0.975 & 0.975 & 0.975 & 0.000 & 0.001 & 0.000 \\
CTU & RLN & Adversarial & tvae & 0.975 & 0.968 & 0.975 & 0.000 & 0.002 & 0.000 \\
CTU & RLN & Adversarial & wgan & 0.975 & 0.974 & 0.975 & 0.000 & 0.001 & 0.000 \\
CTU & RLN & Standard & None & 0.978 & 0.940 & 0.978 & 0.000 & 0.003 & 0.000 \\
CTU & RLN & Standard & ctgan & 0.975 & 0.956 & 0.975 & 0.000 & 0.002 & 0.000 \\
CTU & RLN & Standard & cutmix & 0.953 & 0.953 & 0.953 & 0.000 & 0.000 & 0.000 \\
CTU & RLN & Standard & tablegan & 0.975 & 0.814 & 0.975 & 0.000 & 0.026 & 0.000 \\
CTU & RLN & Standard & tvae & 0.973 & 0.932 & 0.973 & 0.000 & 0.011 & 0.000 \\
CTU & RLN & Standard & wgan & 0.966 & 0.950 & 0.966 & 0.000 & 0.001 & 0.000 \\
CTU & VIME & Adversarial & None & 0.951 & 0.940 & 0.942 & 0.000 & 0.005 & 0.006 \\
CTU & VIME & Adversarial & ctgan & 1.000 & 1.000 & 1.000 & 0.000 & 0.000 & 0.000 \\
CTU & VIME & Adversarial & cutmix & 0.951 & 0.943 & 0.947 & 0.000 & 0.004 & 0.002 \\
CTU & VIME & Adversarial & tablegan & 0.951 & 0.855 & 0.894 & 0.000 & 0.016 & 0.008 \\
CTU & VIME & Adversarial & tvae & 1.000 & 1.000 & 1.000 & 0.000 & 0.000 & 0.000 \\
CTU & VIME & Adversarial & wgan & 0.953 & 0.952 & 0.953 & 0.000 & 0.001 & 0.000 \\
CTU & VIME & Standard & None & 0.951 & 0.408 & 0.951 & 0.000 & 0.049 & 0.000 \\
CTU & VIME & Standard & ctgan & 1.000 & 1.000 & 1.000 & 0.000 & 0.000 & 0.000 \\
CTU & VIME & Standard & cutmix & 0.951 & 0.350 & 0.951 & 0.000 & 0.029 & 0.000 \\
CTU & VIME & Standard & tablegan & 0.951 & 0.670 & 0.951 & 0.000 & 0.021 & 0.000 \\
CTU & VIME & Standard & tvae & 1.000 & 1.000 & 1.000 & 0.000 & 0.000 & 0.000 \\
CTU & VIME & Standard & wgan & 0.953 & 0.229 & 0.953 & 0.000 & 0.022 & 0.000 \\
LCLD & STG & Adversarial & None & 0.156 & 0.121 & 0.156 & 0.000 & 0.001 & 0.000 \\
LCLD & STG & Adversarial & ctgan & 0.820 & 0.812 & 0.820 & 0.000 & 0.001 & 0.000 \\
LCLD & STG & Adversarial & cutmix & 0.376 & 0.362 & 0.376 & 0.000 & 0.000 & 0.000 \\
LCLD & STG & Adversarial & goggle & 0.694 & 0.682 & 0.694 & 0.000 & 0.000 & 0.000 \\
LCLD & STG & Adversarial & tablegan & 0.627 & 0.601 & 0.627 & 0.000 & 0.001 & 0.000 \\
LCLD & STG & Adversarial & tvae & 0.689 & 0.678 & 0.689 & 0.000 & 0.000 & 0.000 \\
LCLD & STG & Adversarial & wgan & 0.613 & 0.597 & 0.613 & 0.000 & 0.000 & 0.000 \\
LCLD & STG & Standard & None & 0.664 & 0.536 & 0.664 & 0.000 & 0.001 & 0.000 \\
LCLD & STG & Standard & ctgan & 0.833 & 0.595 & 0.833 & 0.000 & 0.004 & 0.000 \\
LCLD & STG & Standard & cutmix & 0.352 & 0.222 & 0.352 & 0.000 & 0.002 & 0.000 \\
LCLD & STG & Standard & goggle & 0.577 & 0.433 & 0.577 & 0.000 & 0.002 & 0.000 \\
LCLD & STG & Standard & tablegan & 0.510 & 0.442 & 0.510 & 0.000 & 0.001 & 0.000 \\
LCLD & STG & Standard & tvae & 0.649 & 0.505 & 0.649 & 0.000 & 0.001 & 0.000 \\
LCLD & STG & Standard & wgan & 0.614 & 0.377 & 0.614 & 0.000 & 0.002 & 0.000 \\
LCLD & TabNet & Adversarial & None & 0.000 & 0.000 & 0.001 & 0.000 & 0.000 & 0.000 \\
LCLD & TabNet & Adversarial & ctgan & 0.000 & 0.000 & 0.001 & 0.000 & 0.000 & 0.000 \\
LCLD & TabNet & Adversarial & cutmix & 0.000 & 0.000 & 0.001 & 0.000 & 0.000 & 0.000 \\
LCLD & TabNet & Adversarial & goggle & 1.000 & 1.000 & 1.000 & 0.000 & 0.000 & 0.000 \\
LCLD & TabNet & Adversarial & tablegan & 0.116 & 0.114 & 0.117 & 0.000 & 0.000 & 0.000 \\
LCLD & TabNet & Adversarial & tvae & 0.000 & 0.000 & 0.001 & 0.000 & 0.000 & 0.000 \\
LCLD & TabNet & Adversarial & wgan & 0.000 & 0.000 & 0.001 & 0.000 & 0.000 & 0.000 \\
LCLD & TabNet & Standard & None & 0.674 & 0.004 & 0.674 & 0.000 & 0.001 & 0.000 \\
LCLD & TabNet & Standard & ctgan & 0.029 & 0.021 & 0.030 & 0.000 & 0.001 & 0.000 \\
LCLD & TabNet & Standard & cutmix & 0.000 & 0.000 & 0.001 & 0.000 & 0.000 & 0.000 \\
LCLD & TabNet & Standard & goggle & 0.000 & 0.000 & 0.001 & 0.000 & 0.000 & 0.000 \\
LCLD & TabNet & Standard & tablegan & 0.013 & 0.010 & 0.014 & 0.000 & 0.001 & 0.000 \\
LCLD & TabNet & Standard & tvae & 0.000 & 0.000 & 0.001 & 0.000 & 0.000 & 0.000 \\
LCLD & TabNet & Standard & wgan & 0.000 & 0.000 & 0.001 & 0.000 & 0.000 & 0.000 \\
LCLD & TabTr & Adversarial & None & 0.739 & 0.703 & 0.739 & 0.000 & 0.001 & 0.000 \\
LCLD & TabTr & Adversarial & ctgan & 0.795 & 0.785 & 0.795 & 0.000 & 0.001 & 0.000 \\
LCLD & TabTr & Adversarial & cutmix & 0.725 & 0.710 & 0.725 & 0.000 & 0.001 & 0.000 \\
LCLD & TabTr & Adversarial & goggle & 0.636 & 0.605 & 0.636 & 0.000 & 0.002 & 0.000 \\
LCLD & TabTr & Adversarial & tablegan & 0.608 & 0.564 & 0.608 & 0.000 & 0.003 & 0.000 \\
LCLD & TabTr & Adversarial & tvae & 0.687 & 0.665 & 0.687 & 0.000 & 0.001 & 0.000 \\
LCLD & TabTr & Adversarial & wgan & 0.665 & 0.628 & 0.665 & 0.000 & 0.002 & 0.000 \\
LCLD & TabTr & Standard & None & 0.695 & 0.079 & 0.695 & 0.000 & 0.006 & 0.000 \\
LCLD & TabTr & Standard & ctgan & 0.724 & 0.081 & 0.724 & 0.000 & 0.004 & 0.000 \\
LCLD & TabTr & Standard & cutmix & 0.677 & 0.073 & 0.677 & 0.000 & 0.008 & 0.000 \\
LCLD & TabTr & Standard & goggle & 0.689 & 0.079 & 0.689 & 0.000 & 0.004 & 0.000 \\
LCLD & TabTr & Standard & tablegan & 0.693 & 0.101 & 0.693 & 0.000 & 0.005 & 0.000 \\
LCLD & TabTr & Standard & tvae & 0.703 & 0.048 & 0.703 & 0.000 & 0.003 & 0.000 \\
LCLD & TabTr & Standard & wgan & 0.701 & 0.055 & 0.701 & 0.000 & 0.005 & 0.000 \\
LCLD & RLN & Adversarial & None & 0.695 & 0.630 & 0.695 & 0.000 & 0.001 & 0.000 \\
LCLD & RLN & Adversarial & ctgan & 0.737 & 0.543 & 0.737 & 0.000 & 0.001 & 0.000 \\
LCLD & RLN & Adversarial & cutmix & 0.581 & 0.470 & 0.581 & 0.000 & 0.003 & 0.000 \\
LCLD & RLN & Adversarial & goggle & 0.678 & 0.320 & 0.678 & 0.000 & 0.005 & 0.000 \\
LCLD & RLN & Adversarial & tablegan & 0.688 & 0.479 & 0.688 & 0.000 & 0.004 & 0.000 \\
LCLD & RLN & Adversarial & tvae & 0.670 & 0.643 & 0.670 & 0.000 & 0.000 & 0.000 \\
LCLD & RLN & Adversarial & wgan & 0.661 & 0.402 & 0.661 & 0.000 & 0.004 & 0.000 \\
LCLD & RLN & Standard & None & 0.683 & 0.000 & 0.683 & 0.000 & 0.000 & 0.000 \\
LCLD & RLN & Standard & ctgan & 0.705 & 0.001 & 0.705 & 0.000 & 0.001 & 0.000 \\
LCLD & RLN & Standard & cutmix & 0.689 & 0.000 & 0.689 & 0.000 & 0.000 & 0.000 \\
LCLD & RLN & Standard & goggle & 0.673 & 0.000 & 0.673 & 0.000 & 0.000 & 0.000 \\
LCLD & RLN & Standard & tablegan & 0.693 & 0.001 & 0.693 & 0.000 & 0.001 & 0.000 \\
LCLD & RLN & Standard & tvae & 0.700 & 0.000 & 0.700 & 0.000 & 0.000 & 0.000 \\
LCLD & RLN & Standard & wgan & 0.679 & 0.005 & 0.679 & 0.000 & 0.002 & 0.000 \\
LCLD & VIME & Adversarial & None & 0.655 & 0.104 & 0.655 & 0.000 & 0.002 & 0.000 \\
LCLD & VIME & Adversarial & ctgan & 0.789 & 0.768 & 0.789 & 0.000 & 0.000 & 0.000 \\
LCLD & VIME & Adversarial & cutmix & 0.570 & 0.529 & 0.570 & 0.000 & 0.001 & 0.000 \\
LCLD & VIME & Adversarial & goggle & 0.568 & 0.532 & 0.568 & 0.000 & 0.002 & 0.000 \\
LCLD & VIME & Adversarial & tablegan & 0.563 & 0.537 & 0.563 & 0.000 & 0.000 & 0.000 \\
LCLD & VIME & Adversarial & tvae & 0.678 & 0.661 & 0.678 & 0.000 & 0.001 & 0.000 \\
LCLD & VIME & Adversarial & wgan & 0.617 & 0.530 & 0.617 & 0.000 & 0.002 & 0.000 \\
LCLD & VIME & Standard & None & 0.670 & 0.024 & 0.670 & 0.000 & 0.001 & 0.000 \\
LCLD & VIME & Standard & ctgan & 0.773 & 0.018 & 0.773 & 0.000 & 0.002 & 0.000 \\
LCLD & VIME & Standard & cutmix & 0.523 & 0.020 & 0.523 & 0.000 & 0.001 & 0.000 \\
LCLD & VIME & Standard & goggle & 0.644 & 0.005 & 0.644 & 0.000 & 0.001 & 0.000 \\
LCLD & VIME & Standard & tablegan & 0.607 & 0.005 & 0.607 & 0.000 & 0.001 & 0.000 \\
LCLD & VIME & Standard & tvae & 0.668 & 0.007 & 0.668 & 0.000 & 0.001 & 0.000 \\
LCLD & VIME & Standard & wgan & 0.659 & 0.007 & 0.659 & 0.000 & 0.002 & 0.000 \\
URL & STG & Adversarial & None & 0.943 & 0.900 & 0.903 & 0.000 & 0.001 & 0.001 \\
URL & STG & Adversarial & ctgan & 0.939 & 0.798 & 0.803 & 0.000 & 0.012 & 0.014 \\
URL & STG & Adversarial & cutmix & 0.755 & 0.427 & 0.422 & 0.000 & 0.032 & 0.032 \\
URL & STG & Adversarial & goggle & 0.939 & 0.856 & 0.860 & 0.000 & 0.010 & 0.008 \\
URL & STG & Adversarial & tablegan & 0.921 & 0.809 & 0.816 & 0.000 & 0.004 & 0.003 \\
URL & STG & Adversarial & tvae & 0.957 & 0.795 & 0.804 & 0.000 & 0.017 & 0.015 \\
URL & STG & Adversarial & wgan & 0.942 & 0.812 & 0.813 & 0.000 & 0.003 & 0.003 \\
URL & STG & Standard & None & 0.933 & 0.580 & 0.596 & 0.000 & 0.008 & 0.007 \\
URL & STG & Standard & ctgan & 0.922 & 0.693 & 0.770 & 0.000 & 0.008 & 0.006 \\
URL & STG & Standard & cutmix & 0.794 & 0.397 & 0.444 & 0.000 & 0.009 & 0.010 \\
URL & STG & Standard & goggle & 0.939 & 0.745 & 0.759 & 0.000 & 0.005 & 0.006 \\
URL & STG & Standard & tablegan & 0.876 & 0.469 & 0.575 & 0.000 & 0.005 & 0.008 \\
URL & STG & Standard & tvae & 0.941 & 0.688 & 0.733 & 0.000 & 0.002 & 0.006 \\
URL & STG & Standard & wgan & 0.925 & 0.655 & 0.752 & 0.000 & 0.007 & 0.006 \\
URL & TabNet & Adversarial & None & 0.995 & 0.918 & 0.919 & 0.000 & 0.002 & 0.001 \\
URL & TabNet & Adversarial & ctgan & 0.901 & 0.899 & 0.899 & 0.000 & 0.000 & 0.000 \\
URL & TabNet & Adversarial & cutmix & 0.930 & 0.897 & 0.896 & 0.000 & 0.001 & 0.001 \\
URL & TabNet & Adversarial & goggle & 0.848 & 0.665 & 0.666 & 0.000 & 0.022 & 0.019 \\
URL & TabNet & Adversarial & tablegan & 0.008 & 0.000 & 0.000 & 0.000 & 0.000 & 0.000 \\
URL & TabNet & Adversarial & tvae & 0.940 & 0.872 & 0.870 & 0.000 & 0.018 & 0.018 \\
URL & TabNet & Adversarial & wgan & 0.898 & 0.896 & 0.896 & 0.000 & 0.000 & 0.000 \\
URL & TabNet & Standard & None & 0.934 & 0.110 & 0.299 & 0.000 & 0.005 & 0.004 \\
URL & TabNet & Standard & ctgan & 0.994 & 0.948 & 0.948 & 0.000 & 0.002 & 0.001 \\
URL & TabNet & Standard & cutmix & 0.954 & 0.893 & 0.894 & 0.000 & 0.001 & 0.001 \\
URL & TabNet & Standard & goggle & 0.932 & 0.896 & 0.896 & 0.000 & 0.001 & 0.000 \\
URL & TabNet & Standard & tablegan & 0.896 & 0.878 & 0.875 & 0.000 & 0.010 & 0.011 \\
URL & TabNet & Standard & tvae & 0.938 & 0.891 & 0.892 & 0.000 & 0.002 & 0.003 \\
URL & TabNet & Standard & wgan & 0.998 & 0.952 & 0.953 & 0.000 & 0.002 & 0.001 \\
URL & TabTr & Adversarial & None & 0.939 & 0.567 & 0.578 & 0.000 & 0.009 & 0.009 \\
URL & TabTr & Adversarial & ctgan & 0.930 & 0.660 & 0.664 & 0.000 & 0.004 & 0.004 \\
URL & TabTr & Adversarial & cutmix & 0.850 & 0.403 & 0.404 & 0.000 & 0.011 & 0.012 \\
URL & TabTr & Adversarial & goggle & 0.917 & 0.541 & 0.554 & 0.000 & 0.006 & 0.007 \\
URL & TabTr & Adversarial & tablegan & 0.898 & 0.409 & 0.421 & 0.000 & 0.010 & 0.011 \\
URL & TabTr & Adversarial & tvae & 0.934 & 0.612 & 0.615 & 0.000 & 0.008 & 0.003 \\
URL & TabTr & Adversarial & wgan & 0.927 & 0.569 & 0.580 & 0.000 & 0.008 & 0.010 \\
URL & TabTr & Standard & None & 0.936 & 0.089 & 0.825 & 0.000 & 0.002 & 0.001 \\
URL & TabTr & Standard & ctgan & 0.942 & 0.253 & 0.880 & 0.000 & 0.006 & 0.005 \\
URL & TabTr & Standard & cutmix & 0.904 & 0.018 & 0.687 & 0.000 & 0.000 & 0.000 \\
URL & TabTr & Standard & goggle & 0.930 & 0.049 & 0.051 & 0.000 & 0.001 & 0.001 \\
URL & TabTr & Standard & tablegan & 0.899 & 0.020 & 0.020 & 0.000 & 0.000 & 0.000 \\
URL & TabTr & Standard & tvae & 0.952 & 0.168 & 0.901 & 0.000 & 0.002 & 0.002 \\
URL & TabTr & Standard & wgan & 0.936 & 0.200 & 0.887 & 0.000 & 0.006 & 0.002 \\
URL & RLN & Adversarial & None & 0.952 & 0.562 & 0.566 & 0.000 & 0.007 & 0.006 \\
URL & RLN & Adversarial & ctgan & 0.938 & 0.625 & 0.628 & 0.000 & 0.005 & 0.007 \\
URL & RLN & Adversarial & cutmix & 0.943 & 0.608 & 0.609 & 0.000 & 0.003 & 0.007 \\
URL & RLN & Adversarial & goggle & 0.939 & 0.661 & 0.665 & 0.000 & 0.008 & 0.006 \\
URL & RLN & Adversarial & tablegan & 0.913 & 0.555 & 0.557 & 0.000 & 0.009 & 0.005 \\
URL & RLN & Adversarial & tvae & 0.941 & 0.598 & 0.602 & 0.000 & 0.003 & 0.003 \\
URL & RLN & Adversarial & wgan & 0.933 & 0.547 & 0.552 & 0.000 & 0.002 & 0.005 \\
URL & RLN & Standard & None & 0.944 & 0.108 & 0.901 & 0.000 & 0.002 & 0.001 \\
URL & RLN & Standard & ctgan & 0.942 & 0.219 & 0.855 & 0.000 & 0.005 & 0.001 \\
URL & RLN & Standard & cutmix & 0.941 & 0.086 & 0.926 & 0.000 & 0.002 & 0.002 \\
URL & RLN & Standard & goggle & 0.936 & 0.039 & 0.039 & 0.000 & 0.000 & 0.000 \\
URL & RLN & Standard & tablegan & 0.910 & 0.039 & 0.039 & 0.000 & 0.000 & 0.000 \\
URL & RLN & Standard & tvae & 0.942 & 0.081 & 0.912 & 0.000 & 0.002 & 0.002 \\
URL & RLN & Standard & wgan & 0.935 & 0.214 & 0.911 & 0.000 & 0.002 & 0.002 \\
URL & VIME & Adversarial & None & 0.934 & 0.698 & 0.727 & 0.000 & 0.006 & 0.004 \\
URL & VIME & Adversarial & ctgan & 0.910 & 0.669 & 0.690 & 0.000 & 0.005 & 0.007 \\
URL & VIME & Adversarial & cutmix & 0.920 & 0.686 & 0.707 & 0.000 & 0.010 & 0.012 \\
URL & VIME & Adversarial & goggle & 0.919 & 0.737 & 0.749 & 0.000 & 0.013 & 0.011 \\
URL & VIME & Adversarial & tablegan & 0.887 & 0.645 & 0.652 & 0.000 & 0.005 & 0.004 \\
URL & VIME & Adversarial & tvae & 0.899 & 0.636 & 0.711 & 0.000 & 0.004 & 0.004 \\
URL & VIME & Adversarial & wgan & 0.897 & 0.650 & 0.705 & 0.000 & 0.004 & 0.004 \\
URL & VIME & Standard & None & 0.925 & 0.495 & 0.533 & 0.000 & 0.005 & 0.003 \\
URL & VIME & Standard & ctgan & 0.927 & 0.548 & 0.910 & 0.000 & 0.004 & 0.001 \\
URL & VIME & Standard & cutmix & 0.925 & 0.467 & 0.913 & 0.000 & 0.004 & 0.001 \\
URL & VIME & Standard & goggle & 0.893 & 0.445 & 0.857 & 0.000 & 0.003 & 0.001 \\
URL & VIME & Standard & tablegan & 0.875 & 0.430 & 0.750 & 0.000 & 0.005 & 0.003 \\
URL & VIME & Standard & tvae & 0.909 & 0.444 & 0.886 & 0.000 & 0.005 & 0.003 \\
URL & VIME & Standard & wgan & 0.922 & 0.519 & 0.905 & 0.000 & 0.008 & 0.003 \\
WIDS & STG & Adversarial & None & 0.626 & 0.452 & 0.626 & 0.000 & 0.002 & 0.000 \\
WIDS & STG & Adversarial & ctgan & 0.853 & 0.738 & 0.853 & 0.000 & 0.002 & 0.000 \\
WIDS & STG & Adversarial & cutmix & 0.532 & 0.412 & 0.523 & 0.000 & 0.001 & 0.003 \\
WIDS & STG & Adversarial & goggle & 0.788 & 0.660 & 0.788 & 0.000 & 0.002 & 0.000 \\
WIDS & STG & Adversarial & tablegan & 0.689 & 0.566 & 0.688 & 0.000 & 0.003 & 0.001 \\
WIDS & STG & Adversarial & tvae & 0.677 & 0.598 & 0.677 & 0.000 & 0.001 & 0.001 \\
WIDS & STG & Adversarial & wgan & 0.626 & 0.464 & 0.623 & 0.000 & 0.002 & 0.001 \\
WIDS & STG & Standard & None & 0.776 & 0.638 & 0.773 & 0.000 & 0.002 & 0.000 \\
WIDS & STG & Standard & ctgan & 0.878 & 0.712 & 0.877 & 0.000 & 0.003 & 0.000 \\
WIDS & STG & Standard & cutmix & 0.567 & 0.385 & 0.559 & 0.000 & 0.004 & 0.000 \\
WIDS & STG & Standard & goggle & 0.742 & 0.572 & 0.739 & 0.000 & 0.003 & 0.000 \\
WIDS & STG & Standard & tablegan & 0.677 & 0.498 & 0.671 & 0.000 & 0.004 & 0.000 \\
WIDS & STG & Standard & tvae & 0.759 & 0.621 & 0.755 & 0.000 & 0.003 & 0.000 \\
WIDS & STG & Standard & wgan & 0.746 & 0.583 & 0.744 & 0.000 & 0.002 & 0.000 \\
WIDS & TabNet & Adversarial & None & 0.984 & 0.584 & 0.825 & 0.000 & 0.002 & 0.000 \\
WIDS & TabNet & Adversarial & ctgan & 1.000 & 1.000 & 1.000 & 0.000 & 0.000 & 0.000 \\
WIDS & TabNet & Adversarial & cutmix & 1.000 & 0.374 & 0.671 & 0.000 & 0.003 & 0.007 \\
WIDS & TabNet & Adversarial & goggle & 1.000 & 1.000 & 1.000 & 0.000 & 0.000 & 0.000 \\
WIDS & TabNet & Adversarial & tablegan & 1.000 & 1.000 & 1.000 & 0.000 & 0.000 & 0.000 \\
WIDS & TabNet & Adversarial & tvae & 1.000 & 1.000 & 1.000 & 0.000 & 0.000 & 0.000 \\
WIDS & TabNet & Adversarial & wgan & 1.000 & 0.992 & 0.996 & 0.000 & 0.004 & 0.002 \\
WIDS & TabNet & Standard & None & 0.797 & 0.053 & 0.731 & 0.000 & 0.004 & 0.002 \\
WIDS & TabNet & Standard & ctgan & 1.000 & 1.000 & 1.000 & 0.000 & 0.000 & 0.000 \\
WIDS & TabNet & Standard & cutmix & 0.000 & 0.000 & 0.000 & 0.000 & 0.000 & 0.000 \\
WIDS & TabNet & Standard & goggle & 1.000 & 1.000 & 1.000 & 0.000 & 0.000 & 0.000 \\
WIDS & TabNet & Standard & tablegan & 1.000 & 1.000 & 1.000 & 0.000 & 0.000 & 0.000 \\
WIDS & TabNet & Standard & tvae & 0.984 & 0.406 & 0.475 & 0.000 & 0.001 & 0.000 \\
WIDS & TabNet & Standard & wgan & 0.786 & 0.000 & 0.456 & 0.000 & 0.000 & 0.003 \\
WIDS & TabTr & Adversarial & None & 0.773 & 0.651 & 0.767 & 0.000 & 0.002 & 0.000 \\
WIDS & TabTr & Adversarial & ctgan & 0.781 & 0.681 & 0.776 & 0.000 & 0.003 & 0.001 \\
WIDS & TabTr & Adversarial & cutmix & 0.600 & 0.508 & 0.599 & 0.000 & 0.003 & 0.001 \\
WIDS & TabTr & Adversarial & goggle & 0.765 & 0.675 & 0.755 & 0.000 & 0.002 & 0.001 \\
WIDS & TabTr & Adversarial & tablegan & 0.726 & 0.622 & 0.724 & 0.000 & 0.002 & 0.001 \\
WIDS & TabTr & Adversarial & tvae & 0.747 & 0.667 & 0.743 & 0.000 & 0.002 & 0.001 \\
WIDS & TabTr & Adversarial & wgan & 0.765 & 0.652 & 0.759 & 0.000 & 0.002 & 0.002 \\
WIDS & TabTr & Standard & None & 0.755 & 0.459 & 0.746 & 0.000 & 0.003 & 0.000 \\
WIDS & TabTr & Standard & ctgan & 0.780 & 0.441 & 0.776 & 0.000 & 0.005 & 0.000 \\
WIDS & TabTr & Standard & cutmix & 0.710 & 0.434 & 0.705 & 0.000 & 0.003 & 0.000 \\
WIDS & TabTr & Standard & goggle & 0.786 & 0.383 & 0.733 & 0.000 & 0.004 & 0.000 \\
WIDS & TabTr & Standard & tablegan & 0.750 & 0.376 & 0.750 & 0.000 & 0.008 & 0.000 \\
WIDS & TabTr & Standard & tvae & 0.776 & 0.493 & 0.763 & 0.000 & 0.003 & 0.001 \\
WIDS & TabTr & Standard & wgan & 0.763 & 0.376 & 0.763 & 0.000 & 0.005 & 0.000 \\
WIDS & RLN & Adversarial & None & 0.780 & 0.666 & 0.773 & 0.000 & 0.002 & 0.000 \\
WIDS & RLN & Adversarial & ctgan & 1.000 & 1.000 & 1.000 & 0.000 & 0.000 & 0.000 \\
WIDS & RLN & Adversarial & cutmix & 0.681 & 0.599 & 0.675 & 0.000 & 0.002 & 0.001 \\
WIDS & RLN & Adversarial & goggle & 0.783 & 0.691 & 0.774 & 0.000 & 0.003 & 0.001 \\
WIDS & RLN & Adversarial & tablegan & 0.760 & 0.661 & 0.754 & 0.000 & 0.002 & 0.001 \\
WIDS & RLN & Adversarial & tvae & 0.775 & 0.711 & 0.772 & 0.000 & 0.003 & 0.002 \\
WIDS & RLN & Adversarial & wgan & 0.776 & 0.676 & 0.776 & 0.000 & 0.003 & 0.001 \\
WIDS & RLN & Standard & None & 0.775 & 0.609 & 0.771 & 0.000 & 0.002 & 0.000 \\
WIDS & RLN & Standard & ctgan & 0.762 & 0.472 & 0.759 & 0.000 & 0.007 & 0.000 \\
WIDS & RLN & Standard & cutmix & 0.770 & 0.587 & 0.767 & 0.000 & 0.002 & 0.000 \\
WIDS & RLN & Standard & goggle & 0.773 & 0.525 & 0.750 & 0.000 & 0.001 & 0.000 \\
WIDS & RLN & Standard & tablegan & 0.788 & 0.589 & 0.786 & 0.000 & 0.004 & 0.000 \\
WIDS & RLN & Standard & tvae & 0.802 & 0.621 & 0.796 & 0.000 & 0.004 & 0.000 \\
WIDS & RLN & Standard & wgan & 0.775 & 0.574 & 0.775 & 0.000 & 0.002 & 0.000 \\
WIDS & VIME & Adversarial & None & 0.721 & 0.521 & 0.721 & 0.000 & 0.003 & 0.000 \\
WIDS & VIME & Adversarial & ctgan & 1.000 & 1.000 & 1.000 & 0.000 & 0.000 & 0.000 \\
WIDS & VIME & Adversarial & cutmix & 0.543 & 0.435 & 0.535 & 0.000 & 0.002 & 0.001 \\
WIDS & VIME & Adversarial & goggle & 0.702 & 0.592 & 0.699 & 0.000 & 0.002 & 0.001 \\
WIDS & VIME & Adversarial & tablegan & 0.553 & 0.423 & 0.553 & 0.000 & 0.002 & 0.000 \\
WIDS & VIME & Adversarial & tvae & 0.721 & 0.618 & 0.721 & 0.000 & 0.001 & 0.000 \\
WIDS & VIME & Adversarial & wgan & 0.715 & 0.606 & 0.715 & 0.000 & 0.002 & 0.000 \\
WIDS & VIME & Standard & None & 0.723 & 0.503 & 0.713 & 0.000 & 0.002 & 0.000 \\
WIDS & VIME & Standard & ctgan & 1.000 & 1.000 & 1.000 & 0.000 & 0.000 & 0.000 \\
WIDS & VIME & Standard & cutmix & 0.699 & 0.476 & 0.694 & 0.000 & 0.002 & 0.000 \\
WIDS & VIME & Standard & goggle & 0.702 & 0.491 & 0.697 & 0.000 & 0.003 & 0.000 \\
WIDS & VIME & Standard & tablegan & 0.718 & 0.501 & 0.718 & 0.000 & 0.004 & 0.000 \\
WIDS & VIME & Standard & tvae & 0.726 & 0.506 & 0.726 & 0.000 & 0.004 & 0.000 \\
WIDS & VIME & Standard & wgan & 0.755 & 0.512 & 0.754 & 0.000 & 0.001 & 0.000 \\

\bottomrule

\end{longtable}
}
\clearpage

\subsection{Correlations between ID and robust performances}
\label{subsec:appendix-correlations}

\begin{table}[!h]
    \centering
    
    \caption{Pearson correlations between constrained robust accuracy and: ID accuracy (ID), and non constrained-accuracy (ADV)}
    \label{tab:app-correlations}
\begin{tabular}{ll|rr|rr}
\toprule
Dataset & Training & ID(corr) & ID(p-val) & ADV(corr) & ADV(p-val) \\
\midrule
CTU & Adversarial & 1 & 1.4e-26 & 1 & 1.9e-31 \\
CTU & Standard & 0.22 & 0.28 & 0.22 & 0.28 \\
\midrule
LCLD & Adversarial & 0.76 & 1.8e-06 & 0.76 & 1.8e-06 \\
LCLD & Standard & 0.15 & 0.39 & 0.15 & 0.39 \\
\midrule
URL & Adversarial & 0.7 & 3.6e-06 & 1 & 7.2e-37 \\
URL & Standard & 0.19 & 0.26 & 0.46 & 0.0053 \\
\midrule
WIDS & Adversarial & 0.79 & 1e-06 & 0.91 & 7e-11 \\
WIDS & Standard & 0.031 & 0.87 & 0.62 & 0.00025 \\
\bottomrule
\end{tabular}

\end{table}

\clearpage
\subsection{Impact of budgets, detailed results}
\label{subsec:appendix-budgets}

\begin{figure}[h]
\centering
\begin{subfigure}{0.29\textwidth}
    \includegraphics[clip, trim=0.6cm 1.4cm 0.6cm 0.2cm,width=\textwidth]{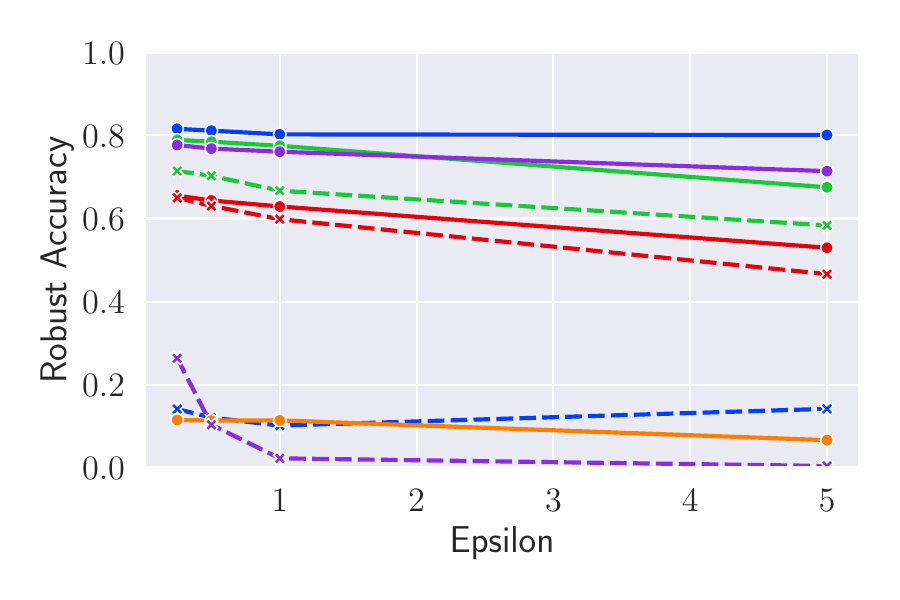}
    \caption{Maximum $\epsilon$ perturbation}
    \label{fig:caa-lcld-eps}
\end{subfigure}
\hfill
\begin{subfigure}{0.29\textwidth}
    \includegraphics[clip, trim=0.6cm 1.4cm 0.6cm 0.2cm,width=\textwidth]{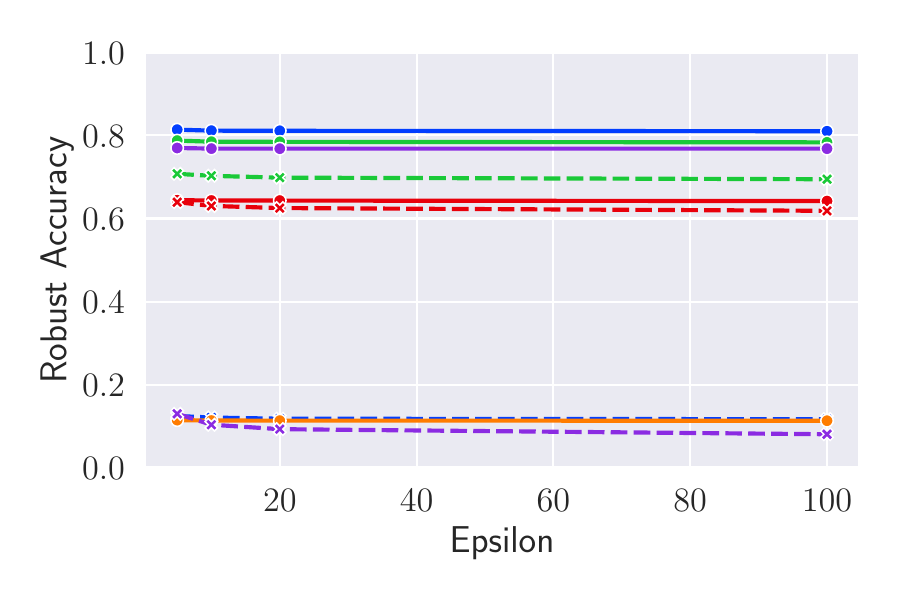}
    \caption{Gradient attack iterations}
    \label{fig:caa-lcld-iter-gradient}
\end{subfigure}
\hfill
\begin{subfigure}{0.29\textwidth}
    \includegraphics[clip, trim=0.6cm 1.4cm 0.6cm 0.0cm,width=\textwidth]{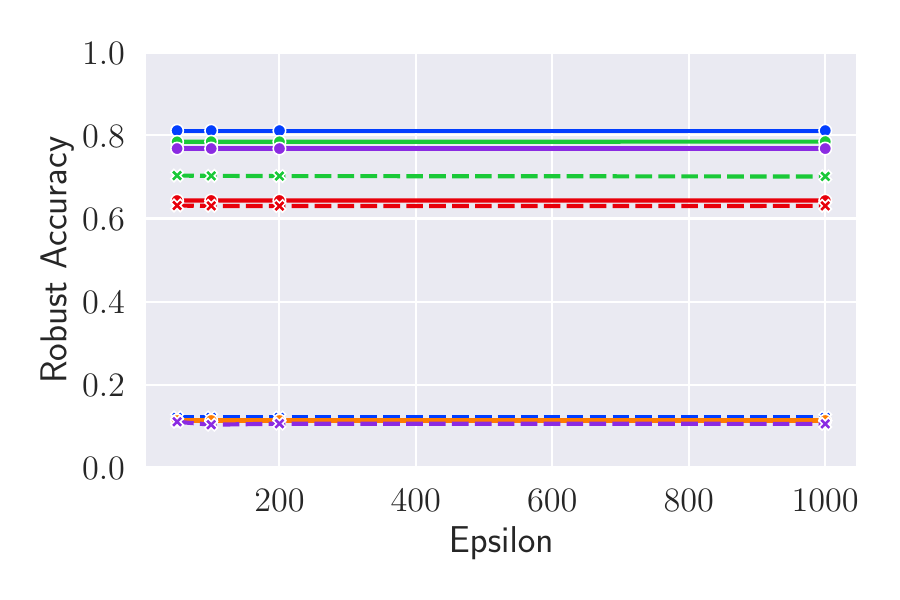}
    \caption{Search attack iterations}
    \label{fig:caa-lcld-iter-search}
\end{subfigure}
\begin{subfigure}{0.10\textwidth}
    \includegraphics[clip, width=\textwidth]{figures/benchmark/legend.pdf}
    \caption*{}
    \label{fig:caa-lcld-legend}
\end{subfigure}
\caption{Impact of attack budget on the robust accuracy for LCLD dataset.}
\label{fig:caa-lcld-budget}
\vspace{-1.5em}
\end{figure}

\begin{figure}[h]
\centering
\begin{subfigure}{0.29\textwidth}
    \includegraphics[clip, trim=0.6cm 1.4cm 0.6cm 0.2cm,width=\textwidth]{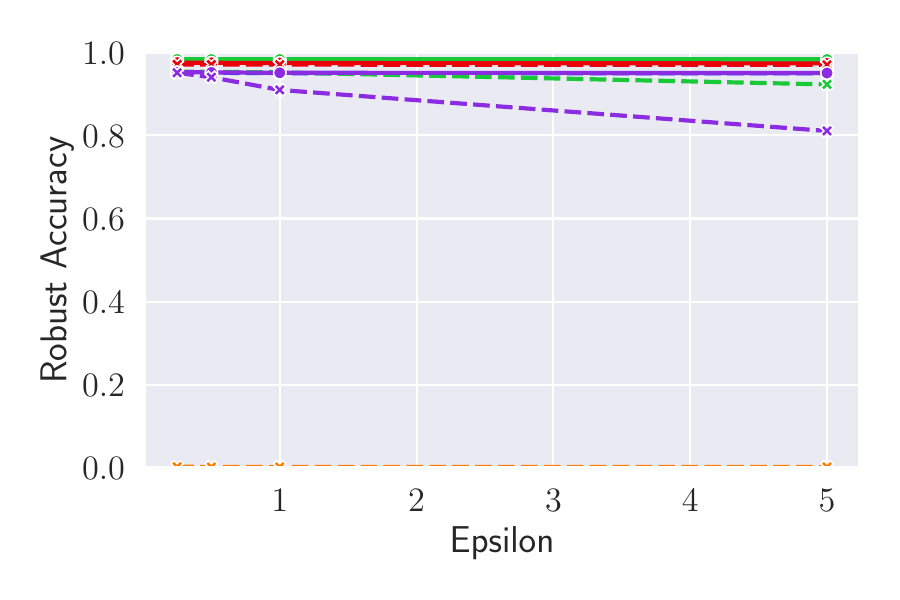}
    \caption{Maximum $\epsilon$ perturbation}
    \label{fig:caa-ctu-eps}
\end{subfigure}
\hfill
\begin{subfigure}{0.29\textwidth}
    \includegraphics[clip, trim=0.6cm 1.4cm 0.6cm 0.2cm,width=\textwidth]{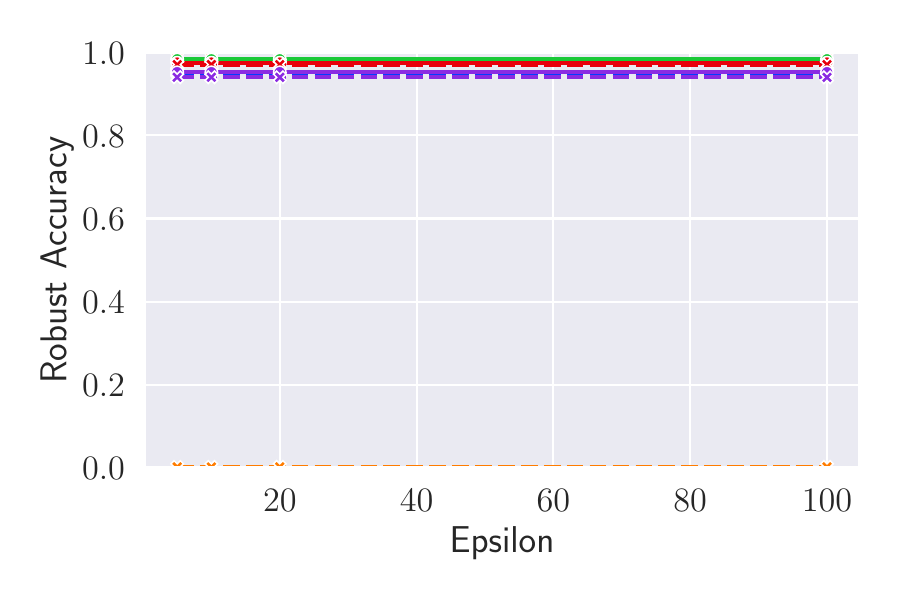}
    \caption{Gradient attack iterations}
    \label{fig:caa-ctu-iter-gradient}
\end{subfigure}
\hfill
\begin{subfigure}{0.29\textwidth}
    \includegraphics[clip, trim=0.6cm 1.4cm 0.6cm 0.0cm,width=\textwidth]{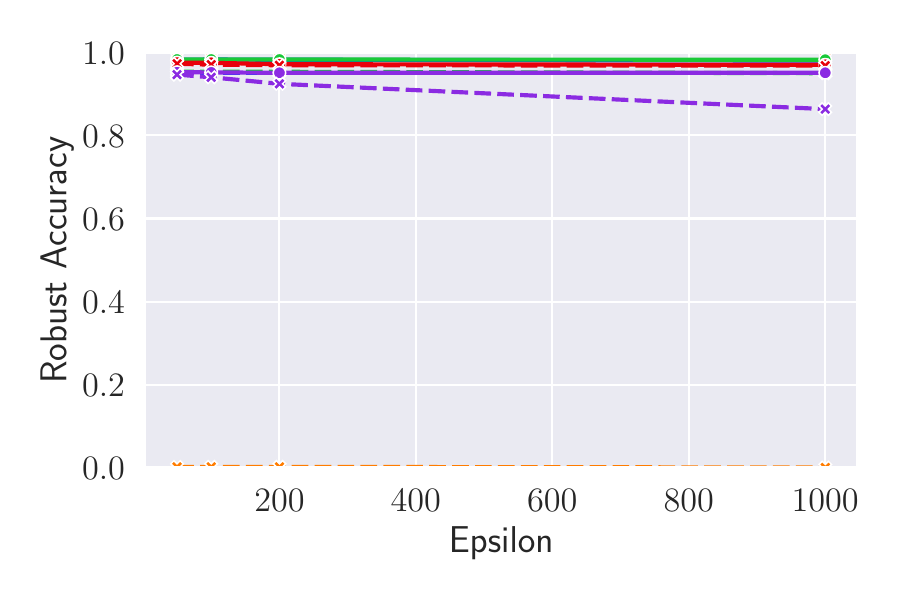}
    \caption{Search attack iterations}
    \label{fig:caa-ctu-iter-search}
\end{subfigure}
\begin{subfigure}{0.10\textwidth}
    \includegraphics[clip, width=\textwidth]{figures/benchmark/legend.pdf}
    \caption*{}
    \label{fig:caa-ctu-legend}
\end{subfigure}
\caption{Impact of attack budget on the robust accuracy for CTU dataset.}
\label{fig:caa-ctu-budget}
\vspace{-1.5em}
\end{figure}

\begin{figure}[h]
\centering
\begin{subfigure}{0.29\textwidth}
    \includegraphics[clip, trim=0.6cm 1.4cm 0.6cm 0.2cm,width=\textwidth]{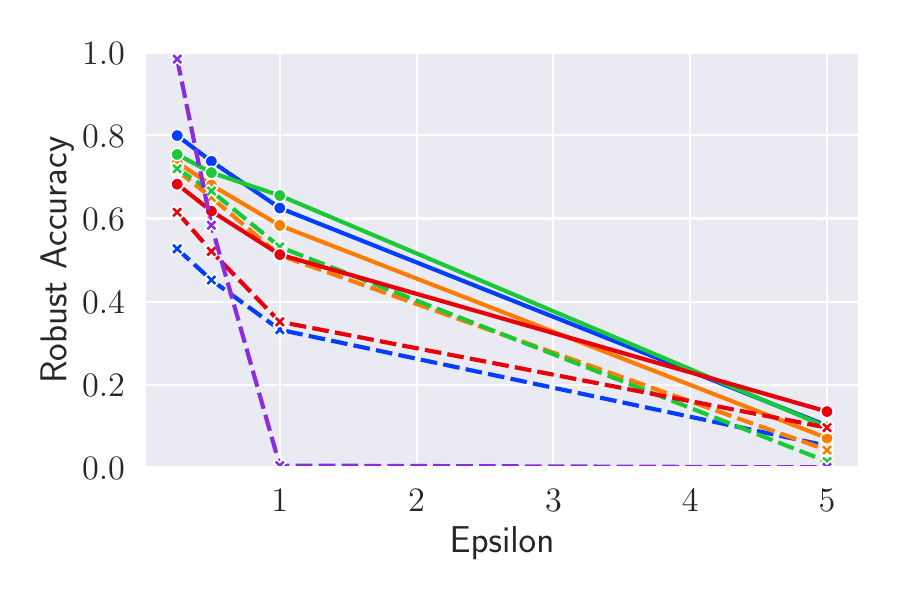}
    \caption{Maximum $\epsilon$ perturbation}
    \label{fig:caa-wids-eps}
\end{subfigure}
\hfill
\begin{subfigure}{0.29\textwidth}
    \includegraphics[clip, trim=0.6cm 1.4cm 0.6cm 0.2cm,width=\textwidth]{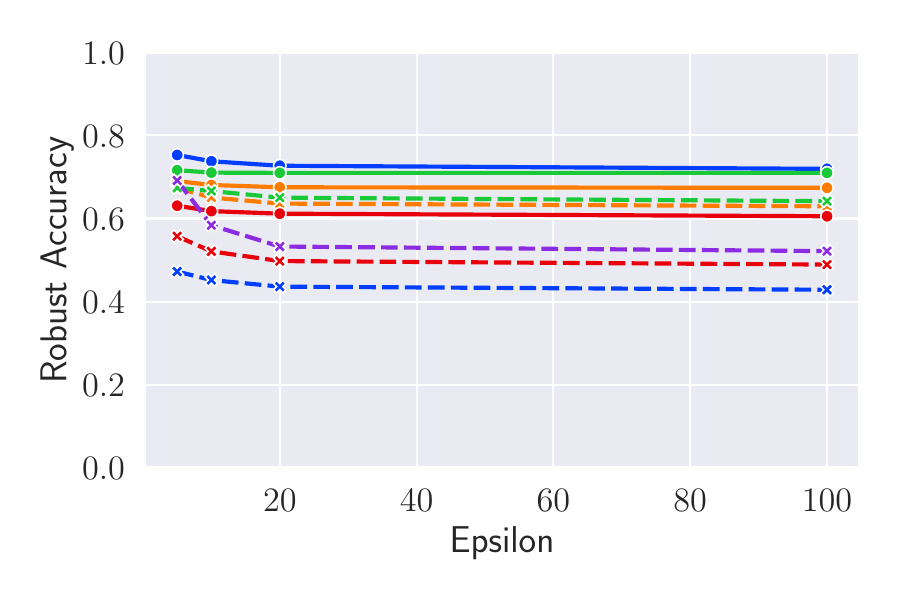}
    \caption{Gradient attack iterations}
    \label{fig:caa-wids-iter-gradient}
\end{subfigure}
\hfill
\begin{subfigure}{0.29\textwidth}
    \includegraphics[clip, trim=0.6cm 1.4cm 0.6cm 0.0cm,width=\textwidth]{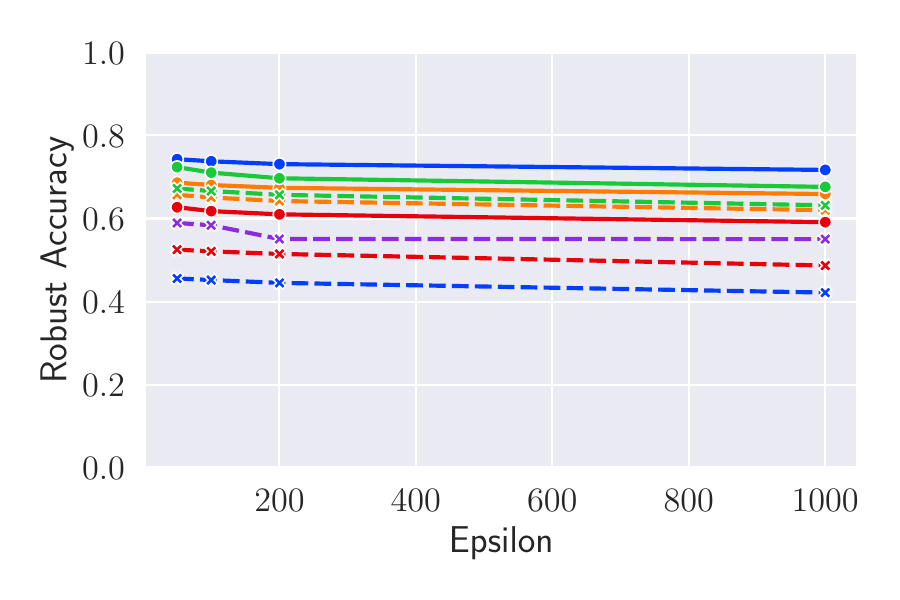}
    \caption{Search attack iterations}
    \label{fig:caa-wids-iter-search}
\end{subfigure}
\begin{subfigure}{0.10\textwidth}
    \includegraphics[clip, width=\textwidth]{figures/benchmark/legend.pdf}
    \caption*{}
    \label{fig:caa-wids-legend}
\end{subfigure}
\caption{Impact of attack budget on the robust accuracy for WIDS dataset.}
\label{fig:caa-wids-budget}
\vspace{-1.5em}
\end{figure}

\begin{figure}[h]
\centering
\begin{subfigure}{0.29\textwidth}
    \includegraphics[clip, trim=0.6cm 1.4cm 0.6cm 0.2cm,width=\textwidth]{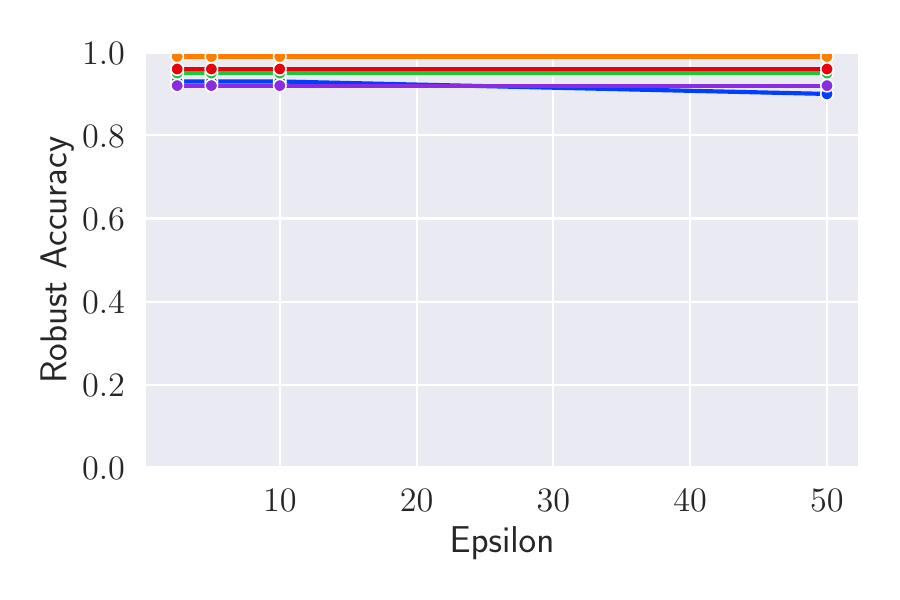}
    \caption{Maximum $\epsilon$ perturbation}
    \label{fig:caa-wids-eps}
\end{subfigure}
\hfill
\begin{subfigure}{0.29\textwidth}
    \includegraphics[clip, trim=0.6cm 1.4cm 0.6cm 0.2cm,width=\textwidth]{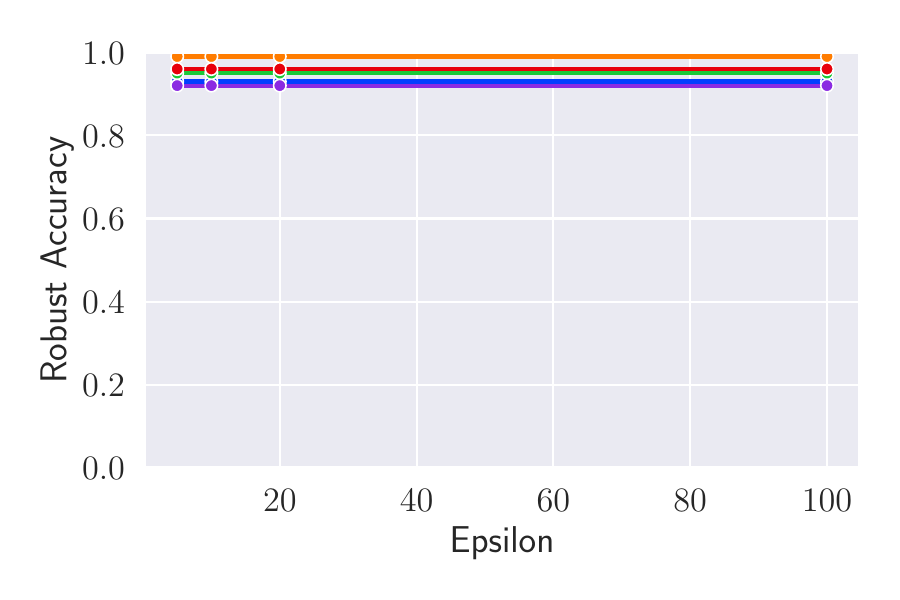}
    \caption{Gradient attack iterations}
    \label{fig:caa-wids-iter-gradient}
\end{subfigure}
\hfill
\begin{subfigure}{0.29\textwidth}
    \includegraphics[clip, trim=0.6cm 1.4cm 0.6cm 0.0cm,width=\textwidth]{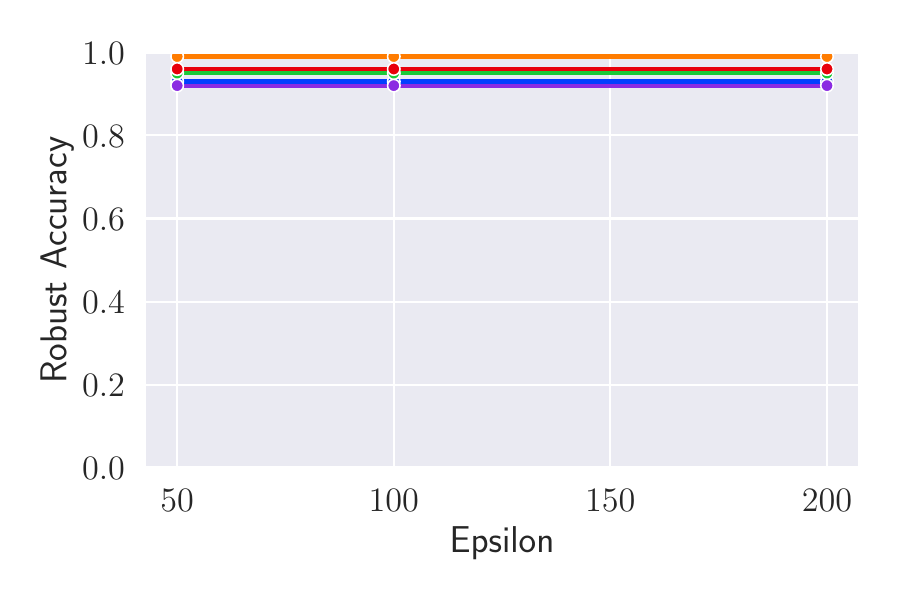}
    \caption{Search attack iterations}
    \label{fig:caa-wids-iter-search}
\end{subfigure}
\begin{subfigure}{0.10\textwidth}
    \includegraphics[clip, width=\textwidth]{figures/benchmark/legend.pdf}
    \caption*{}
    \label{fig:caa-wids-legend}
\end{subfigure}
\caption{Impact of attack budget on the robust accuracy for Malware dataset.}
\label{fig:caa-wids-budget}
\vspace{-1.5em}
\end{figure}

\clearpage

\subsection{Generalization to other distances}
\label{subsec:appendix-distance}

We define for all attacks a distance function. This method is used for MOEVA (the evolution attack) to measure the fitness value related to the distance objective, and in the evaluation method to validate the correctness of the adversarial examples.

By default, it supports $L_\infty$ and $L_2$ distances \footnote{\url{https://github.com/serval-uni-lu/tabularbench/blob/main/tabularbench/attacks/utils.py}}:

\begin{python}
from tabularbench.utils.typing import NDBool, NDInt, NDNumber

def compute_distance(x_1: NDNumber, x_2: NDNumber, norm: Any) -> NDNumber:
    if norm in ["inf", np.inf, "Linf", "linf"]:
        distance = np.linalg.norm(x_1 - x_2, ord=np.inf, axis=-1)
    elif norm in ["2", 2, "L2", "l2"]:
        distance = np.linalg.norm(x_1 - x_2, ord=2, axis=-1)
    else:
        raise NotImplementedError

    return distance

\end{python}

One can define any new distance metric, like structural similarity index measure (SSIM), or some semantic measure after embedding the features $x_1$ and $x_2$. The distance used here does not need to be differentiable and is not backpropagated in the gradient attacks.

Hence, for CAPGD component of the benchmark attack, we need to define a custom project mechanism for each distance.
We implemented a projection over sphere of $L_\infty$ and $L_2$ distances \url{https://github.com/serval-uni-lu/tabularbench/blob/main/tabularbench/attacks/capgd/capgd.py#L196}.

To extend the projected gradient attacks to other distances, custom projection mechanisms are then needed.

\section{API}
\label{sec:appendx_api}

The library \url{https://github.com/serval-uni-lu/tabularbench/tree/main/tabularbench} is split in 4 main components. The \emph{test} folder provides meaningful examples for each component.

\subsection{Datasets}

Our dataset factory support 5 datasets: CTU, LCLD, MALWARE, URL, and WIDS.
each dataset can be invoked with the following aliases:

\begin{python}
from tabularbench.datasets import dataset_factory

dataset_aliases= [
        "ctu_13_neris",
        "lcld_time",
        "malware",
        "url",
        "wids",
    ]
    
for dataset_name in dataset_aliases:
    dataset = dataset_factory.get_dataset(dataset_name)
    x, _ = dataset.get_x_y()
    metadata = dataset.get_metadata(only_x=True)
    assert x.shape[1] == metadata.shape[0]

\end{python}

Each dataset can be defined in a single .py file (example: \url{https://github.com/serval-uni-lu/tabularbench/blob/main/tabularbench/datasets/samples/url.py}).

A dataset needs at least a source (local or remote csv) for the raw features, and a definition of feature constraints. The said definition can be empty for non-constrained datasets.

\subsection{Constraints}

One of the features of our benchmark is the support of feature constraints, but in the dataset definition and in the attacks. 

Constraints can be expressed in natural language. For example, we express the constraint $F_0 = F_1 + F_2$ such as:

\begin{python}
from tabularbench.constraints.relation_constraint import Feature
constraint1 = Feature(0) == Feature(1) + Feature(2)
\end{python}

Given a dataset, one can check the constraint satisfcation over all constraints, given a tolerance.

\begin{python}
from tabularbench.constraints.constraints_checker import ConstraintChecker
from tabularbench.datasets import dataset_factory

dataset = dataset_factory.get_dataset("url")
x, _ = dataset.get_x_y()

constraints_checker = ConstraintChecker(
    dataset.get_constraints(), tolerance
)
out = constraints_checker.check_constraints(x.to_numpy())

\end{python}

In the provided datasets, all constraints are satisfied. During the attack, Constraints can be fixed as follows:

\begin{python}
import numpy as np
from tabularbench.constraints.constraints_fixer import ConstraintsFixer

x = np.arange(9).reshape(3, 3)

constraints_fixer = ConstraintsFixer(
            guard_constraints=[constraint1],
            fix_constraints=[constraint1],
        )

x_fixed = constraints_fixer.fix(x)

x_expected = np.array([[3, 1, 2], [9, 4, 5], [15, 7, 8]])

assert np.equal(x_fixed, x_expected).all()

\end{python}

Constraint violations can be translated into losses and one can compute the gradient to repair the faulty constraints as follows:

\begin{python}
import torch

from tabularbench.constraints.constraints_backend_executor import (
    ConstraintsExecutor,
)

from tabularbench.constraints.pytorch_backend import PytorchBackend
from tabularbench.datasets.dataset_factory import get_dataset

ds = get_dataset("url")
constraints = ds.get_constraints()
constraint1 = constraints.relation_constraints[0]

x, y = ds.get_x_y()
x_metadata = ds.get_metadata(only_x=True)
x = torch.tensor(x.values, dtype=torch.float32)

constraints_executor = ConstraintsExecutor(
    constraint1,
    PytorchBackend(),
    feature_names=x_metadata["feature"].to_list(),
)
        
x.requires_grad = True
loss = constraints_executor.execute(x)
grad = torch.autograd.grad(
    loss.sum(),
    x_l,
)[0]

\end{python}

\subsection{Models}

All models need to extend the class \textbf{BaseModelTorch}\footnote{\url{https://github.com/serval-uni-lu/tabularbench/blob/main/tabularbench/models/torch_models.py}}
. This class implements the definitions, the fit and evaluation methods, and the save and loading methods. Depending of the architectures, scaler and feature encoders can be required by the constructors.

So far, our API nateively supports: multi-layer preceptrons (MLP), RLN, STG, TabNet, TabTransformer, and VIME. Our implementation is based on Tabsurvey \cite{borisov2021deep}. All models from this framework can be easily adapted to our API.  

\subsection{Benchmark}

The leaderboard is available on \url{https://serval-uni-lu.github.io/tabularbench/}.

This leaderboard will be updated regularly, and all the models listed in leaderboard are downloadable using our API

\begin{figure}[!ht]
    \centering
    \includegraphics[width=\linewidth]{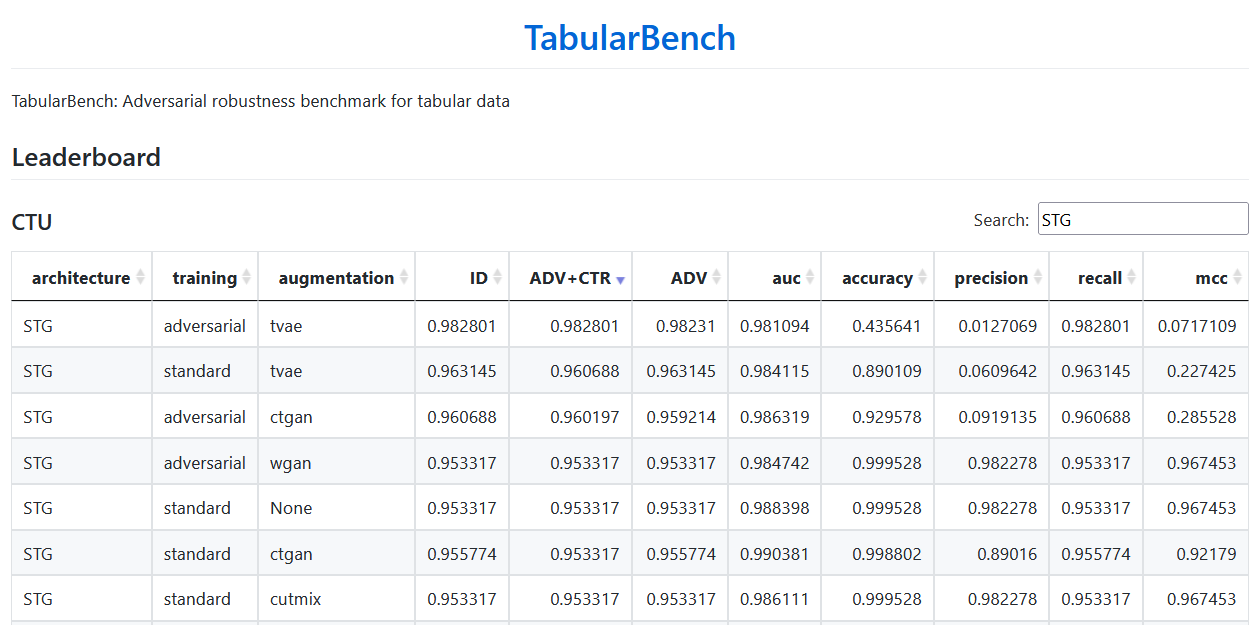}
    \caption{Screenshot of the TabularBench leaderboard on 12/06/2024}
    \label{fig:app-leaderboard}
\end{figure}

The benchmark leverages Constrained Adaptive Attack (CAA) by default and can be extended for other attacks.

\begin{python}
clean_acc, robust_acc = benchmark(dataset='LCLD', model="TabTr_Cutmix", distance='L2', constraints=True)
\end{python}

The model attribute refers to a pre-trained model in the relevant model folder. The API infers the architecture from the first term of the model name, but it can be defined manually. In the above example, a \textbf{TabTransformer} architecture will be initialized.

\end{document}